\documentclass{article}

    \usepackage[final]{neurips_2021}

\usepackage[utf8]{inputenc} 
\usepackage[T1]{fontenc}    
\usepackage{hyperref}       
\usepackage{url}            
\usepackage{booktabs}       
\usepackage{amsfonts}       
\usepackage{nicefrac}       
\usepackage{microtype}      
\usepackage{xcolor}         
\usepackage{times,latexsym}
\usepackage{url}
\usepackage{times}
\usepackage{array}
\usepackage{tikz}
\usepackage{soul}
\usepackage{caption, floatrow}

\usepackage{times}
\usepackage{graphicx}
\usepackage{multirow}
\usepackage{amssymb}  
\usepackage{amsthm}
\usepackage{adjustbox}
\usepackage{amsmath}
\usepackage[official]{eurosym}
\usepackage{mathtools}
\newcommand{\Tref}[1]{Table~\ref{#1}}
\newcommand{\Eref}[1]{Eq.~(\ref{#1})}
\newcommand{\Fref}[1]{Fig.~\ref{#1}}
\newcommand{\Aref}[1]{Alg.~\ref{#1}}
\newcommand{\Sref}[1]{\S~\ref{#1}}
\usepackage{arydshln}
\usepackage[linesnumbered,ruled,vlined]{algorithm2e}

\SetCommentSty{mycommfont}
\usepackage{graphicx}
\usepackage{calrsfs}
\usepackage{array}
\usepackage{xcolor}
\usepackage[T1]{fontenc}
\usepackage{CJKutf8}
\usepackage[OT2, T1]{fontenc}
\usepackage[russian, english]{babel}
\usepackage[T1]{fontenc}
\usepackage{fdsymbol}
\usepackage[ruled,vlined]{algorithm2e}
\usepackage{wrapfig,lipsum}
\usepackage{paralist}
\newcommand{\tydi}{\textsc{TyDi~QA}\xspace}

\newcommand{\tydixor}{\textsc{Xor}-\textsc{TyDi} QA\xspace}

\newcommand{\xorfull}{\textsc{Xor-Full}\xspace}
\newcommand{\mora}{CORA\xspace}
\newcommand{\cora}{CORA\xspace}
\DeclareMathOperator{\bert}{mBERT}
\SetKwFunction{Fn}{Function}

\definecolor{darkgreen}{HTML}{228B22}

\usepackage{subfig}
\usepackage{wrapfig}
\usepackage{lipsum}
\usepackage{caption}
\usepackage{color,soul}
\newfloatcommand{capbtabbox}{table}[][0.25\linewidth]
\usepackage[utf8]{inputenc}
\usepackage{CJKutf8}

\title{One Question Answering Model for Many Languages with Cross-lingual Dense Passage Retrieval}

\author{
  Akari Asai$^{\dagger}$, Xinyan Yu$^{\dagger}$, Jungo Kasai$^{\dagger}$, Hannaneh Hajishirzi$^{{\dagger}{\ddagger}}$ \\
  $^{\dagger}$University of Washington, $^{\ddagger}$Allen Institute for AI\\
  \texttt{\{akari, xyu530, jkasai, hannaneh\}@cs.washington.edu} \\
}
\begin{document}

\maketitle
\begin{abstract}
We present {\bf C}ross-lingual {\bf O}pen-{\bf R}etrieval {\bf A}nswer Generation ({\bf \mora}), the first unified many-to-many question answering (QA) model that can answer questions across many languages, even for ones without language-specific annotated data or knowledge sources.
We introduce a new dense passage retrieval algorithm that is trained to retrieve documents across languages for a question.
Combined with a  multilingual autoregressive generation model, \mora answers directly in the target language without any translation or in-language retrieval modules as used in prior work. 
We propose an iterative training method that automatically extends annotated data available only in high-resource languages to low-resource ones.
Our  results show that \mora substantially outperforms the previous state of the art on multilingual open QA benchmarks across 26 languages, 9 of which are unseen during training. 
Our  analyses show the significance of cross-lingual retrieval and generation in many languages, particularly under low-resource settings. 
Our code and trained model are publicly available at \url{https://github.com/AkariAsai/CORA}.
\end{abstract}
\section{Introduction}
Multilingual open question answering (QA) is the task of answering a question from a large collection of multilingual documents. Most recent progress in  open QA is made for English by building a pipeline based on a dense passage retriever trained on large-scale English QA datasets to find evidence passages in English ~\citep{lee-chang-toutanova:2019:ACL2019,karpukhin2020dense}, followed by a reader that extracts an answer from retrieved passages. 
However, extending this approach to  multilingual open QA poses new challenges. 
Answering multilingual questions requires retrieving evidence from knowledge sources of other languages than the original question since many languages have limited reference documents or the question sometimes inquires about concepts from other cultures \citep{xorqa,lin2020pretrained}. 
Nonetheless, large-scale cross-lingual open QA training data whose questions and evidence are in different languages are not available in many of those languages.

To address these challenges, previous work in multilingual open QA~\citep{ture-boschee-2016-learning,xorqa} translates questions into English, applies an English open QA  system to answer in English, and then translates answers back to the target language. 
Those pipeline approaches suffer from error propagation of the machine translation component into the downstream QA, especially for low-resource languages. Moreover, they are not able to answer questions whose answers can be found in resources written in languages other than English or the target languages. 

\begin{figure*}[t!]
  \includegraphics[width=\linewidth]{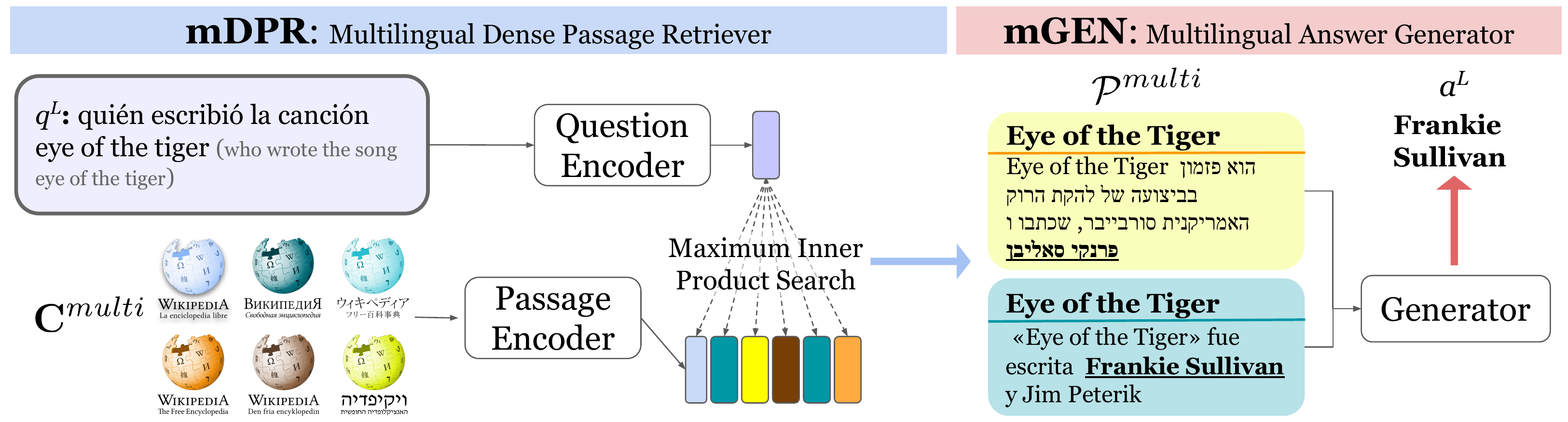}
  \caption{Overview of \mora (mDPR and mGEN).}
  \label{img:method_overview}
\end{figure*}

In this paper, we introduce a unified \textit{many-to-many} QA model that can answer questions in \textit{any} target language by retrieving evidence from \textit{any} language and generating answers in the target language.
Our method (called \mora, \Fref{img:method_overview}) extends the {\it retrieve-then-generate} approach of English open QA~\citep{lewis2020retrieval,izacard2020leveraging} with a single cross-lingual retriever and a generator that do not rely on language-specific retrievers or machine translation modules. 
The multilingual retrieval module ({\bf mDPR}) produces dense embeddings of a question and all multilingual passages, thereby retrieving passages across languages.
The generation module ({\bf mGEN}) is trained to output an answer in the target language conditioned on the retrieved multilingual passages. 
To overcome the aforementioned data scarcity issue, we automatically mine training data using external language links and train mDPR and mGEN iteratively.
In particular, each iteration proceeds over two stages of updating model parameters with available training data and mining new training data {\it cross-lingually} by Wikipedia language links and predictions made by the models.
This approach does not require any additional human annotations or machine translation, and can be applied to many new languages with low resources. 

Our experiments show that \mora advances the state of the art on two multilingual open QA datasets, \tydixor \citep{xorqa} and MKQA \citep{mkqa}, across 26 typologically diverse languages; \mora achieves gains of 23.4 and 4.7 F1 points in \tydixor and MKQA respectively, where MKQA data is not used for training. 
Moreover, \mora achieves F1 scores of roughly 30 over 8 languages on MKQA that have no training data or even reference Wikipedia documents, outperforming the state-of-the-art approach by 5.4 F1 points. 
Our controlled experiments and human analyses illustrate the impact of many-to-many cross-lingual retrieval in improving multilingual open QA performance.
We further observe that through cross-lingual retrieval, \cora can find answers to 20\% of the multilingual questions that are valid but are originally annotated as {\it unanswerable} by humans due to the lack of evidence in the English knowledge sources.

\section{Method}
\label{sec:overview}
We define multilingual open QA as the task of answering a question $q^L$ in a target language $L$ given a collection of multilingual reference passages $\mathbf{C}^{multi}$, where evidence passages can be retrieved from \textit{any language}. 
These passages come from Wikipedia articles that are not necessarily parallel over languages. 
We introduce \mora, which runs a \textit{retrieve-then-generate} procedure to achieve this goal (\Fref{img:method_overview}).
We further introduce a novel training scheme of iterative training with data mining (\Sref{subsec:iterative_training}).

\subsection{\mora Inference}
\mora directly retrieves evidence passages from \textit{any} language for questions asked in \textit{any} target language, and then generates answers in the target language conditioned on those passages.
More formally, the \mora inference consists of two steps of (i) retrieving passages $\mathcal{P}^{multi}$ and (ii) generating an answer $a^L$ based on the retrieved passages. 
$\mathcal{P}^{multi}$ can be in any language included in $\mathbf{C}^{multi}$.
\begin{equation*}
\mathcal{P}^{multi} = \text{mDPR}(q^L, \mathbf{C}^{multi} ),~ 
  a^L = \text{mGEN}(q^L, \mathcal{P}^{multi} ). 
\end{equation*}

\noindent{\bf Multilingual Dense Passage Retriever (mDPR).}
mDPR extends Dense Passage Retriever (DPR; \citealp{karpukhin2020dense}) to a multilingual setting.
mDPR uses an iterative training approach to fine-tune a pre-trained multilingual language model (e.g., mBERT;~\citealp{devlin2018bert}) to encode passages and questions separately. 
Once training is done, the representations for all passages from $\mathbf{C}^{multi}$ are computed offline and stored locally. 
Formally, a passage encoding is obtained as follows: $\mathbf{e}_{p^L} = \bert_{p}(p)$, where a passage $p$ is a fixed-length sequence of tokens from multilingual documents.
At inference, mDPR independently obtains a $d$-dimensional ($d=768$)  encoding of the question $\mathbf{e}_{q^L} = \bert_{q}(q^L)$.
It  retrieves $k$ passages with the $k$ highest relevance scores to the question, where the relevance score between a passage $p$ and a question $q^L$ is estimated by the inner product of their encoding vectors, $\langle \mathbf{e}_{q^L}, \mathbf{e}_p \rangle$.

\noindent{\bf Multilingual Answer Generator (mGEN). }
\label{sec:method_reader}
We use a multilingual sequence-to-sequence model (e.g., mT5;~\citealp{xue2020mt5}) to generate answers in the target language token-by-token given the retrieved multilingual passages $\mathcal{P}^{multi}$. 
We choose a generation approach because it can generate an answer in the target language $L$ from passages across different languages.\footnote{An alternative approach of answer extraction requires translation for all language pairs \citep{xorqa}.}
Moreover, the generator can be adapted to unseen languages, some of which may have little or no translation training data. 
Specifically, the generator outputs the sequence probability for $a^L$ as follows:
\begin{equation}
  P(a^L | q^L, \mathcal{P}^{multi}) = \prod_{i}^T p(a^L_{i} | a^L_{<i}, q^L, \mathcal{P}^{multi}),
  \label{eq:generator}
\end{equation}
where $a^L_i$ denotes the $i$-th token in the answer, and $T$ is the length of the answer.
We append a language tag to the question to indicate the target language.

\subsection{\mora Training}
\label{subsec:iterative_training}
\begin{figure*}[t!]
\vspace{-0.5em}
  \includegraphics[width=\linewidth]{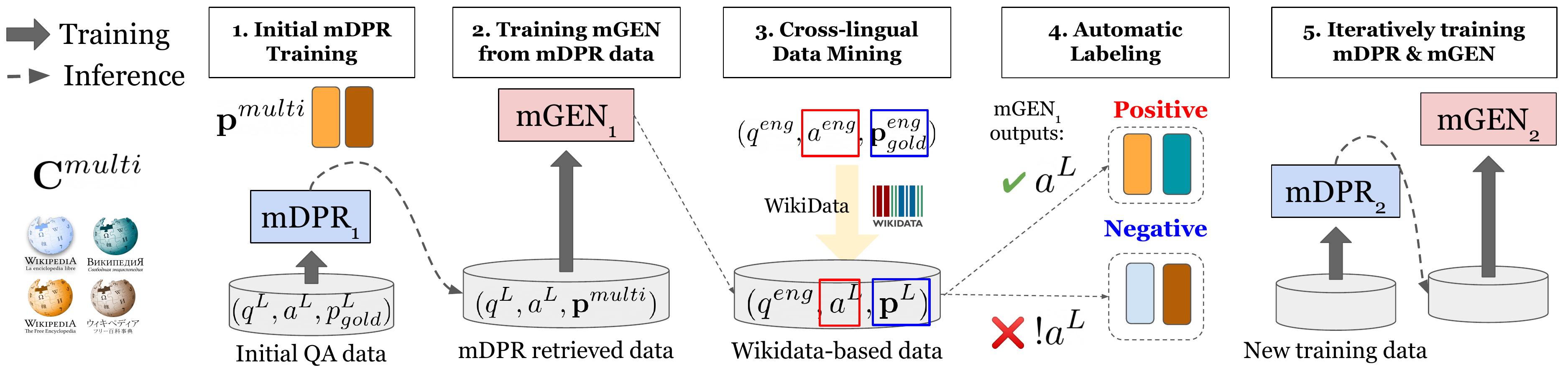}
  \caption{Overview of \cora iterative training and data mining. 
}
\vspace{-0.5em}
  \label{img:overview_training}
\end{figure*}
We introduce an iterative training approach that encourages cross-lingual retrieval and answer generation conditioned on multilingual passages (sketched in \Fref{img:overview_training} and \Aref{algo:training}).
Each iteration proceeds over two stages: \textbf{parameter updates} (\Sref{sec:param_updates}) where mDPR and mGEN are trained on the current training data and \textbf{cross-lingual data mining} (\Sref{sec:data_mining}) where training data are automatically expanded by Wikipedia language links and model predictions. 

\vspace{-.3cm} \paragraph{Initial training data.}
The initial training data is a combination of multilingual QA datasets: \tydixor and \tydi~\citep{tydiqa}, and an English open QA dataset (Natural Questions, \citealp{kwiatkowski2019natural}). 
Each training instance from these datasets comprises a question, a positive passage, and an answer.
Note that annotations in the existing QA datasets have critical limitations: positive passages are taken either from English~\citep{xorqa} or the question's language~\citep{tydiqa}.
Further, most of the non-English languages are not covered.
Indeed, when we only train mDPR on this initial set, it often learns to retrieve passages in the same languages or similar languages with irrelevant context or context without sufficient evidence to answer.

\subsubsection{Parameter Updates}
\label{sec:param_updates}
\textbf{mDPR updates} (line 3 in \Aref{algo:training}).
Let $\mathcal{D} = \{ \langle q_i^L, p^+_i, p^-_{i,1}, \cdots, p^-_{i,n} \rangle \}_{i=1}^m$ be $m$ training instances. Each instance consists of a question $q_i^L$, a passage that answers the question (positive passage) $p^+_i$, and $n$ passages that do not answer the question (negative passages) $p^-_{i,j}$. 
For each question, we use positive passages for the other questions in the training batch as negative passages (\textit{in-batch negative}, ~\citealp{gillick-etal-2019-learning,karpukhin2020dense}).
mDPR is updated by minimizing the negative log likelihood of positive passages:
\begin{equation}
  \mathcal{L}_{\text{mdpr}} = -\log \frac{\exp(\langle \mathbf{e}_{q_i^L},\, \mathbf{e}_{p_i^+}\rangle)}{\exp(\langle \mathbf{e}_{q^L_i},\, \mathbf{e}_{p_i^+}\rangle) + \sum_{j=1}^n{\exp(\langle \mathbf{e}_{q^L_i},\, \mathbf{e}_{p^-_{i,j}}\rangle)}}.
  \label{eq:nll-loss}
\end{equation}
\textbf{mGEN updates} (lines 4-5 in \Aref{algo:training}).
After updating mDPR, we use mDPR to retrieve top $k$ passages $\mathcal{P}^{multi}$ for each $q^L$. 
Given these pairs of the question and the retrieved passages $(q^L, \mathcal{P}^{multi})$ as input, mGEN is trained to generate answer $a^L$ autoregressively (\Eref{eq:generator}) and minimize the cross-entropy loss. 
To train the model to generate in languages not covered by the original datasets, we translate $a^L$ to other languages using Wikipedia language links and create new synthetic answers.\footnote{This automatic answer translation is only done after the third epoch of initial training to prevent the model from overfitting to synthetic data. }
See Appendix \Sref{app_sec:data_mining} for more detail.

\subsubsection{Cross-lingual Data Mining}
\label{sec:data_mining}
\setcounter{algocf}{0}
\begin{algorithm}[t!]
\captionsetup{labelfont={sc,bf}}
\caption{Iterative training that automatically mines training data.}
\label{algo:training}
\footnotesize
\SetAlgoLined
\KwData{Input QA pairs: $(q^L, a^L)$}
 initialize training data $\mathbf{B}^1=(q^\mathbf{L}, a^\mathbf{L}, p_{gold}), \mathbf{L} = \{{\rm Eng}, L\}$\;
 \While{$t < T$}{
 $\Theta_{mDPR}^t \gets Train(\theta_{mDPR}^{t-1}, \mathbf{B}^t )$\tcc{Train mDPR}
 $\mathcal{P}^{multi} \gets {\rm mDPR}(q^\mathbf{L}, \mbox{embedding}(\mathbf{C}^{multi})) $\tcc{Retrieve passages}
 $\theta_{mGEN}^t \gets Train(\theta_{mGEN}^{t-1}, (q^\mathbf{L}, a^\mathbf{L}, \mathcal{P}^{multi}))$\tcc{Train mGEN}
  \mbox{For} {$\mathbf{L} == {\rm Eng}$,} $\mathcal{P}^{multi} +=  {\rm LangLink}(q^{\mathbf{L}}, \mathbf{C}^{multi})) $ \tcc{Mine data using Wikidata }
\mbox{\bf For} {$p_i \in \mathcal{P}^{multi}$:} 
  \mbox{\bf if} {{\rm mGEN}$(q^\mathbf{L}, p_i) == a^L$} \mbox{\bf then} $positives.add(p_i)$\ 
  \mbox{\bf else } $negatives.add(p_i)$\\
 $\mathbf{B}^{t+1}$ +=  $(q^\mathbf{L}, a^\mathbf{L}, positives,negatives) $ \tcc{Add new training  data}
 $t \gets t +1 $
}
\end{algorithm}
After the parameter updates, we mine new training data using mDPR and Wikipedia language links and label the new data by mGEN predictions.
This step is skipped in the final iteration.

\textbf{Mining by trained mDPR and language links} (line 4, 6 in \Aref{algo:training}).
Trained mDPR can discover positive passages in another language that is not covered by the initial training data. 
At each iteration, we use retrieved passages $\mathcal{P}^{multi}$ for $q^L$ (line 4 in \Aref{algo:training}) as a source of new positive and negative passages. 
This enables expanding data between language pairs not in the original data. 

To cover even more diverse languages, we use language links and find passages in other languages that potentially include sufficient evidence to answer. 
Wikipedia maintains article-level language links that connect articles on the same entity over languages. 
We use these links to expand training data from the English QA dataset of Natural Questions (line 6 in \Aref{algo:training}).
Denote a training instance by $(q^{En}, a^{En}, p_{gold})$.
We first translate the English answer $a^{En}$ to a target language $a^L$ using language links.
We use language links again to look up the English Wikipedia article that the gold passage $p_{gold}$ comes from.
We then find articles in non-English languages in the reference documents $\mathbf{C}^{multi}$ that correspond to this article. 
Although the language link-based automatic translation cannot handle non-entity answers (e.g., short phrases), this helps us to scale to new languages without additional human annotation or machine translation. 
We add all passages from these articles to $\mathcal{P}^{multi}$ as positive passage candidates, which are then passed to mGEN to evaluate whether each of them leads to $a^L$ or not. 

\textbf{Automatic labeling by mGEN predictions} (lines 7-8 in \Aref{algo:training}). 
A passage $p_i$ from $\mathcal{P}^{multi}$ may not always provide sufficient information to answer the question $q^L$ even when it includes the answer string $a^L$.
To filter out those {\it spurious passages}~\citep{lin-etal-2018-denoising,min2019discrete}, we take instances generated from the two mining methods described above, and run mGEN on each passage to predict an answer for the question. 
If the answer matches the correct answer $a^L$, then the passage $p_i$ is labeled as a {\it positive  passage}; otherwise we label the input passage as a {\it negative passage}. 
We assume that when mGEN fails to generate a correct answer given the passage, the passage may not provide sufficient evidence to answer; this helps us filter out spurious passages that accidentally contain an answer string yet do not provide any clue to answer.
We add these new positive and negative passages to the training data, and in the next iteration, mDPR is trained on this expanded training set (\Sref{sec:param_updates}).

\section{Experiments}
\label{sec:experiments}
We evaluate \mora on two multilingual open QA datasets across 28 typologically diverse languages.\footnote{A full list of the language families and script types are in the appendix.}
\mora achieves state-of-the-art performance across 26 languages, and greatly outperforms previous approaches that use language-specific components such as question or answer translation.

\subsection{Datasets and Knowledge Sources}
Multilingual open QA datasets differ in covered languages, annotation schemes, and target application scenarios. 
We evaluate F1 and EM scores over the questions with answer annotations from two datasets, following the common evaluation practice in open QA~\citep{lee-chang-toutanova:2019:ACL2019}.

\noindent {\bf \tydixor.}
\tydixor~\citep{xorqa} is a multilingual open QA dataset consisting of 7 typologically diverse languages, where questions are originally from \tydi~\citep{tydiqa} and posed by information-seeking native speakers.
The answers are annotated by extracting spans from Wikipedia in the same language as the question ({\it in-language data}) or by translating English spans extracted from English Wikipedia to the target language ({\it cross-lingual data}). \tydixor offers both training and evaluation data.

\noindent {\bf MKQA.}
MKQA~\citep{mkqa} is an evaluation dataset created by translating 10k Natural Questions~\citep{kwiatkowski2019natural} to 25 target languages.
The parallel data enables us to compare the models' performance across typologically diverse languages, in contrast to \tydixor. 
MKQA has evaluation data only; \tydixor and MKQA have five languages in common.

\vspace{.1cm}
\noindent {\bf Collection of multilingual documents $\mathbf{C}^{multi}$.} 
We use the February 2019 Wikipedia dumps of 13 diverse languages from all \tydixor languages and a subset of MKQA languages.\footnote{ {Downloaded from \url{ https://archive.org/details/wikimediadownloads?and\%5B\%5D=year\%3A\%222019\%22}.}} 
We choose 13 languages to cover languages with a large number of Wikipedia articles and a variety of both Latin and non-Latin scripts. 
We extract plain text from Wikipedia articles using wikiextractor,\footnote{\url{https://github.com/attardi/wikiextractor}} and split each article into 100-token segments as in DPR~\citep{karpukhin2020dense}. 
We filter out disambiguation pages that distinguish pages that share the same article title\footnote{\url{https://en.wikipedia.org/wiki/Category:Disambiguation_pages}.}
as well as pages with fewer than 20 tokens, resulting in 43.6M passages.
See more details in Appendix  \Sref{app_sec:wikipedia_stat}.

\noindent {\bf Language categories.} 
To better understand the model performance, we categorize the languages based on their availability during our training.
We call the languages with human annotated gold paragraph and answer data {\it seen} languages. \tydixor provides gold passages for 7 languages. 
For the languages in $\mathbf{C}^{multi}$ without human-annotated passages, we mine new mDPR training data by our iterative approach. We call these languages, which are seen during mDPR training, {\it mDPR-seen}. 
We also synthetically create mGEN training data as explained in \Sref{sec:param_updates} by simply replacing answer entities with the corresponding ones in the target languages.  
The languages that are unseen by mDPR but are seen by mGEN {\it mGEN-seen}, and all other languages (i.e., included neither in mDPR nor mGEN training) {\it unseen languages}.
9 of the MKQA languages are unseen languages.

\subsection{Baselines and Experimental Setting}
\label{sec:baselines}
We compare \mora with the following strong baselines adopted from \citet{xorqa}.

\noindent {\bf Translate-test (MT + DPR).} 
As used in most previous work (e.g., \citealp{xorqa}), this method translates the question to English, extracts an answer in English using DPR, and then translates the answer back to the  target language. The translation models are obtained from MarianMT~\citep{mariannmt} and trained on the OPUS-MT dataset~\citep{tiedemann-2012-parallel}.

\noindent {\bf Monolingual baseline (BM25).}
This baseline retrieves passages solely from the target language and extracts the answer from the retrieved passages. 
Training neural network models such as DPR is infeasible with a few thousands of training examples.
Due to the lack of training data in most of the target languages, we use a BM25-based lexical retriever implementation by Pyserini~\citep{lin2021pyserini}. 
We then feed the retrieved documents to a multilingual QA model to extract final answers.

\noindent {\bf MT+Mono.}
This baseline combines results from the  translate-test method and the monolingual method to retrieve passages in both English and the target language.
Following \cite{xorqa}, we prioritize predictions from the monolingual pipeline if they are over a certain threshold tuned on \tydixor development set; otherwise we output predictions from the translate-test method.\footnote{For the languages not supported by Pyserini, we always output translate-test's predictions.}

\noindent {\bf Closed-book baseline.}
This model uses an mT5-base\footnote{We did not use larger-sized variants due to our computational budget.}  sequence-to-sequence model that  takes a question as input and generates an answer in the target language without any retrieval at inference time~\citep{roberts2020much}.
This baseline assesses the models' ability to memorize and retrieve knowledge from its parameters without retrieving reference documents. 

\noindent {\bf \mora details.}  For all experiments, we use a single retriever (mDPR) and a single generator (mGEN) that use the same passage embeddings.
mDPR uses multilingual BERT base uncased,\footnote{The alternative of XLM-RoBERTa~\citep{conneau-etal-2020-unsupervised} did not improve our results.} and the generator fine-tunes mT5-base.
We found that using other pre-trained language models such as mBART~\citep{liu-etal-2020-multilingual-denoising} for mGEN or XLM-R~\citep{conneau-etal-2020-unsupervised} for mDPR did not improve performance and sometimes even hurt performance. 
We first fine-tune mDPR using gold passages from Natural Questions, and then further fine-tune it using \tydixor and \tydi's gold passage data. 
We exclude the training questions in Natural Questions and \tydi that were used to create the MKQA or \tydixor evaluation set.
We run two iterations of \cora training (\Sref{subsec:iterative_training}) after the initial fine-tuning.
All hyperparameters are in Appendix~\Sref{app_sec:hyper_params}.

\section{Results and Analysis}
\label{sec:results}
\subsection{Multilingual Open QA Results}
\paragraph{\tydixor.}
\Tref{table:main_results} reports the scores of \mora and the baselines in \tydixor.  
\mora, which only uses a single retriever and a single generator, outperforms the baselines and the previous state-of-the-art model on \tydixor by a large margin across all 7 languages. 
\mora achieves gains of 24.8 macro-averaged F1 points over the previous state-of-the-art method (GMT+GS), which uses external black-box APIs, and 23.4 points over the concurrent anonymous work (SER). 

\begin{table*}[t!]
\addtolength{\tabcolsep}{-0.5pt}
\small
    \centering
    \begin{tabular}{l|ccccccc|ccc}
\toprule
 Models & \multicolumn{7}{c}{Target Language $L_i$ F1} & \multicolumn{3}{|c}{Macro Average} \\
 & {\bf Ar} & {\bf Bn} & {\bf Fi} & {\bf Ja} & {\bf Ko} & {\bf Ru} & {\bf Te} & {\bf F1} & {\bf EM} & {\bf BLEU}  \\ \midrule
\mora & {\bf 59.8} &  {\bf 40.4} & {\bf 42.2} & {\bf 44.5} & {\bf 27.1} & {\bf 45.9} & {\bf 44.7} & \textbf{43.5} & {\bf 33.5} &  {\bf 31.1} \\
SER &  32.0 & 23.1 & 23.6 & 14.4 & 13.6 & 11.8 & 22.0 & 20.1 & 13.5 & 20.1  \\
GMT+GS &  31.5 & 19.0 & 18.3 & 8.8 & 20.1 & 19.8 & 13.6 & 18.7 & 12.1 & 16.8  \\
MT+Mono & 25.1  & 12.7 & 20.4  & 12.9 & 10.5 & 15.7 & 0.8 & 14.0 & 10.5 &  11.4 \\
MT+DPR &  7.6 & 5.9 & 16.2 & 9.0 & 5.3 & 5.5 & 0.8 & 7.2 & 3.3 & 6.3 \\
BM25 & 31.1 & 21.9 & 21.4 & 12.4 & 12.1 & 17.7 & -- & -- & -- & -- \\
\hdashline
Closed-book & 14.9 & 10.0 & 11.4 & 22.2 & 9.4 & 18.1 & 10.4 & 13.8 & 9.6 & 7.4 \\
 \bottomrule
    \end{tabular}
    \caption{Performance on \xorfull (test data F1 scores and macro-averaged F1, EM and BLEU scores). 
    ``GMT+GS'' denotes the previous state-of-the-art model, which combines Google Custom Search in the target language and Google Translate + English DPR for cross-lingual retrieval \citep{xorqa}. Concurrent to our work, ``SER'' is a state-of-the-art model, Single Encoder Retriever, submitted anonymously on July 14 to the \xorfull leaderboard (\url{https://nlp.cs.washington.edu/xorqa/}). 
    We were not able to find a BM25 implementation that supports Telugu.
    }
    \label{table:main_results}
\end{table*}

\noindent {\bf MKQA.}
Tables \ref{table:main_results_mkqa_grouped_seen} and \ref{table:main_results_mkqa_grouped_unseen} report the F1 scores of \mora and the baselines on over 6.7k MKQA questions with short answer annotations\footnote{
Following previous work in open QA but different from the official script of MKQA \citep{mkqa}, we disregard the questions labeled as ``no answer''.
As shown in our human analysis, it is
difficult to prove an answer does not exist in the millions of multilingual documents even if the annotation says so.
} under {\it seen} and {\it unseen} settings.
\mora significantly outperforms the baselines in all languages by a large margin except for Arabic and English. 
Note that \citet{mkqa} report results in a simplified setting with gold reference articles from the original Natural Questions dataset given in advance, and thus their results are not comparable.
\mora yields larger improvements over the translate-test baseline in the languages that are distant from English and with limited training data such as Malay (Ms; 27.8 vs.\ 12.6) and Hebrew (He; 15.8 vs.\ 8.9). 
The performance drop of the translate-test model from English (43.3 F1) to other languages indicates the error propagation from the translation process.
BM25 performs very poorly in some low-resource languages such as Thai because of the lack of answer content in the target languages' Wikipedia.
MT+Mono underpeforms the MT+DPR baseline in MKQA since it is challenging to rerank answers from two separate methods with uncaliberated confidence scores.
In contrast, \mora retrieves passages across languages, achieving around 30 F1 on a majority of the 26 languages.

\begin{table*}[t!]
\footnotesize
    \centering
   \begin{tabular}{l||ccccccccccc}
\toprule
  Setting & -- & \multicolumn{6}{c}{Seen (Included in \tydixor)  } & \multicolumn{4}{|c}{ mDPR-seen } \\\hline
 & Avg. over all $L$. & {\bf En} &  {\bf Ar} & {\bf Fi} & {\bf Ja} & {\bf Ko} & {\bf Ru}  & {\bf Es} & {\bf Sv} & {\bf He} & {\bf Th} \\\hline
 \midrule
\mora  &  {\bf 21.8}  & 40.6   &  12.8 & {\bf 26.8} & {\bf 19.7} & {\bf 12.0} &{\bf 19.8} &{\bf 32.0} &  {\bf 30.9} & {\bf 15.8} & {\bf 8.5}  \\
MT+Mono&  14.1  & 19.3  &  6.9 & 17.5 &  9.0 & 7.0 & 10.6 & 21.3 & 20.0 & 8.9 & 8.3 \\
MT+DPR&  17.1  & {\bf 43.3} & {\bf 16.0} & 21.7 & 9.6 & 5.7 & 17.6 & 28.4 & 19.7 & 8.9 & 6.9 \\
BM25 &  -- & 19.4 & 5.9 & 9.9 & 9.1 & 6.9 & 8.1 & 14.7 & 10.9 & -- &  4.9 \\\hdashline
Closed& 4.5  & 8.0 & 4.6 & 3.6 & 6.5 & 3.8 & 4.1 & 6.6 & 4.8 & 3.8 & 2.1  \\ 
\bottomrule
    \end{tabular}
    \caption{F1 scores on MKQA seen and mDPR-seen languages. 
    }
    \label{table:main_results_mkqa_grouped_seen}
\end{table*}
\begin{table*}[t!]
\scriptsize
    \centering
 \hspace{-10mm}
   \begin{tabular}{l||ccccccccccccccccc}
\toprule
 Setting & \multicolumn{7}{c}{ mGEN-seen } & \multicolumn{9}{|c}{ Unseen}\\\hline
&  {\bf Da} & {\bf De} &  {\bf Fr}  & {\bf It}  & {\bf Nl}  & {\bf Pl}  & {\bf Pt}   &  {\bf Hu}  & {\bf Vi}   & {\bf Ms}  &   {\bf Km} & {\bf No} & {\bf Tr} & {\bf cn} & {\bf hk} &  {\bf tw} \\ \midrule
 \midrule
\mora  & {\bf 30.4} & {\bf 30.2} & {\bf 30.8} & {\bf 29.0} & {\bf 32.1} & {\bf 25.6} & {\bf 28.4} & {\bf 18.4} & {\bf 20.9} & {\bf 27.8} & {\bf 5.8} & {\bf 29.2} & {\bf 22.2} & {\bf 5.2} & {\bf 6.7} & {\bf 5.4} \\
MT+Mono& 19.3 & 21.6 & 21.9 & 20.9 & 21.5 & 24.6 & 19.9 &  16.5 & 15.1 &  12.6 &  1.2 & 17.4 & 16.6 &  4.9 &  3.8 & 5.1  \\
MT+DPR&  26.2 &  25.9 & 21.9 & 25.1 & 28.3 & 24.6 & 24.7 & 15.7 & 15.1 & 12.6 & 1.2 & 18.3 &  18.2 & 3.3 & 3.8 & 3.8 \\
BM25 &  9.5 & 12.5 & -- & 13.6 & 12.8 & -- & 13.4 & 7.4 & -- & -- & -- & 9.4 & 8.8 & 2.8 & -- & 3.3 \\\hdashline
Closed& 4.7 & 5.6 & 5.8 & 5.3 &  5.5 & 4.0 & 4.4 & 5.5 &  5.9 &  5.3 & 1.9 & 4.1 & 3.8 & 2.6 & 2.3 & 2.4 \\ 
\bottomrule
    \end{tabular}
    \caption{F1 scores on MKQA in mGEN-seen and unseen languages.}``cn'': ``Zh-cn'' (Chinese, simplified). ``hk'': ``Zh-hk'' (Chinese, Hong Kong). ``tw'':``Zh-tw'' (Chinese, traditional). 
    \label{table:main_results_mkqa_grouped_unseen}
\end{table*}
\subsection{Analysis}
\label{sec:analysis}

\paragraph{Ablations: Impact of \mora components.}
\begin{table}[ht!]
\centering
\small
\addtolength{\tabcolsep}{-1.2pt}
\begin{tabular}{l|cccc|cccccc}
\toprule
\textbf{Setting} & \multicolumn{4}{c|}{\tydixor} &\multicolumn{6}{c}{MKQA} \\
 & Avg. F1 & Ar & Ja & Te   & Avg. F1 & Fi & Ru & Es & Th & Vi \\
\midrule
\mora & {\bf 31.4} & {\bf 42.6} &  {\bf 33.4} & {\bf 26.1 }& {\bf 22.3 }& {\bf 25.9} &  {\bf 20.6}& {\bf 33.2} & 6.3 & {\bf 22.6} \\ 
(i) mDPR$_1$ + mGEN$_1$ & 27.9  &36.2 & 29.8 & 21.1   & 17.3 &  23.1  & 13.1 & 28.5 & 5.7 & 18.6 \\
(ii) DPR (trained NQ)+mGEN & 24.3 & 30.7 & 29.2 & 19.0 & 17.9  & 20.1  &  16.9 & 29.4 & 5.5  & 18.2 \\
(iii) \mora, $\mathbf{C}^{multi}$=\{En\} & 19.1  & 20.5 &  23.2 &  11.5 & 20.5 & 24.7  & 15.4 & 28.3 & {\bf 8.3}  &21.9 \\
(iv) mDPR+Ext.reader+MT &  11.2 & 11.8 & 10.8 & 5.6 & 12.2 & 16.1 & 10.9 & 25.2 & 1.2 & 12.7 \\
\bottomrule
\end{tabular}
\caption{Ablation studies on \tydixor development set and a subset of  MKQA. 
}
\label{table:ablation}
 \end{table}
We compare \mora with the following four variants to study the impact of different components.
(i) {\bf mDPR$_1$ + mGEN$_1$} only trains \mora using the initial labeled, annotated data and measures the impact of the iterative training.
(ii) {\bf DPR (trained NQ) + mGEN} replaces mDPR with a multilingual BERT-based DPR trained on English data from Natural Questions (NQ), and encodes all passages in $\mathbf{C}^{multi}$. 
This configuration assesses the impact of cross-lingual training data. 
(iii) {\bf \mora, $\mathbf{C}^{multi}$=\{En\} } only retrieves from English during inference. 
This variant evaluates if English reference documents suffice to answer multilingual questions. 
(iv) {\bf mDPR+Ext.reader+MT} replaces mGEN with an extractive reader model followed by answer translation. This variant quantifies the effectiveness of using a multilingual generation model over the approach that combines an extractive reader model with language-specific translation models. 
Note that for MKQA experiments, we sample the same 350 questions ($\sim$5\%) from the evaluation set for each language to reduce the computational cost over varying configurations. 

Results in \Tref{table:ablation} show performance drops in all variants.
This supports the following claims: (i) the iterative learning and data mining process is useful, (ii) mDPR trained with cross-lingual data substantially outperforms DPR with multilingual BERT, (iii) reference languages other than English are important in answering multilingual questions, and (iv) a multilingual generation model substantially boosts the model performance.

\vspace{-.2cm}\paragraph{Retrieval performance and relationship to the final QA performance.}
 \begin{table}[ht!]
\centering
\small
\addtolength{\tabcolsep}{-1.6pt}
\begin{tabular}{l|l||ccccc|cccccc}
\toprule
 & Setting  & \multicolumn{5}{c}{{\bf mDPR-Seen}}| & \multicolumn{6}{c}{{\bf Unseen}} \\\hline
 & Lang  & Es & Fi & Ja & Ru & Th & Pt &  Ms & Tr & Zh-Cn & Zh-Hk & Km \\\hdashline
& Script  & \multicolumn{2}{c}{Latn}| & Jpan |& Cyrl | & Thai &  \multicolumn{3}{c}{Latn} | &\multicolumn{2}{c}{Hant} | &Khmr \\\hline
\midrule
\multirow{2}{*}{mDPR}& {\scriptsize \texttt{R$^\texttt{L}$@10} }& 53.7  & 52.8 & 32.9 &  42.3 &14.9 & 50.0 & 49.4 &  42.0 &  12.6 &  16.6 & 15.7 \\
& {\scriptsize \texttt{R$^\texttt{multi}$@10}} & 63.4  &   60.9 & 42.0   &  54.0 & 28.0&  62.6 &  63.4 & 55.4 & 40.6 & 42.3 & 25.1 \\\hdashline
DPR(NQ) &{\scriptsize \texttt{R$^\texttt{L}$@10} } & 52.3 & 46.0  & 24.6  & 36.0 & 12.6 & 45.7 & 48.8 & 32.0  & 9.1 & 14.0 & 13.4 \\ 
 &{\scriptsize \texttt{R$^\texttt{multi}$@10}} &  63.1 &  53.1 & 32.9 & 49.1 & 29.4 & 56.8 &58.0  & 44.0 & 36.3 & 39.4 & 23.4 \\
\bottomrule
\end{tabular}
\caption{
Retrieval recall performance on MKQA as the percentage of the questions where at least one out of the top 10 passages includes an answer string in the target language (\texttt{R$^\texttt{L}$@10}), or in any language  (\texttt{R$^\texttt{multi}$@10}). The same subset of the MKQA evaluation data are used as in the ablations. 
}
\label{table:retrieval}
 \end{table}
We evaluate \mora's retrieval performance on MKQA using two recall metrics that measure the percentage of questions with at least one passage among the top $10$ that includes a string in an answer set in the target language (\texttt{R$^\texttt{L}$@10}) or in the union of answer sets from all languages that are available in MKQA (\texttt{R$^\texttt{multi}$@10}).
MKQA provides answer translations across 26 languages.

\Tref{table:retrieval} reports retrieval results for {\bf mDPR} and multilingual BERT-based DPR trained on NQ: {\bf DPR (NQ)}.
This is equivalent to (ii) from the ablations. 
We observe that mDPR performs well in Indo-European languages with Latin script, even when the language is unseen. 
Interestingly, there is a significant performance gap between \texttt{R$^\texttt{L}$@10} and \texttt{R$^\texttt{multi}$@10} in languages with non-Latin script (e.g., Japanese, Russian, Chinese); this suggests that our model often uses relevant passages from other languages with Latin script such as English or Spanish to answer questions in those languages with non-Latin script.
Our mDPR outperforms DPR (NQ) by a large margin in unseen languages with limited resources, which are consistent with the findings in \Tref{table:main_results_mkqa_grouped_unseen}.
Nevertheless, we still see low performance on Khmer and Thai even with the \texttt{R$^\texttt{multi}$@10} metric.
We also observe that passage and query embeddings for those languages are far from other languages, which can be further studied in future work.
We provide a two-dimensional visualization of the encoded passage representations in the appendix.

\begin{figure}
\begin{floatrow}
    \begin{minipage}{0.68\textwidth}
        \ffigbox[.98\textwidth]{
        \includegraphics[width=\textwidth]{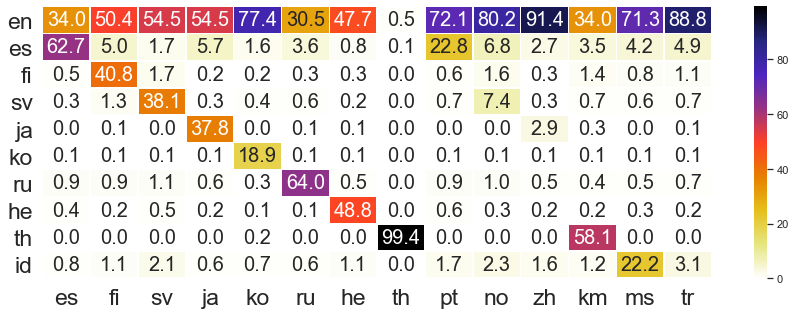}\vspace{-.7cm}
        }{
        \caption{
              Breakdown of the languages of retrieved reference passages for sampled MKQA questions (\%). The x and y axes indicate target (question) and retrieval reference languages respectively.\label{fig:heatmap}
            }
        }
    \end{minipage}
    \begin{minipage}{0.3\textwidth}
        \setlength{\tabcolsep}{2.5pt}
        \ffigbox[0.99\textwidth]{}{
            \adjustbox{}{
                \begin{tabular}{l|cc}
                    \toprule
                     & Ja &  Es  \\
                    \midrule
                    retrieval errors & 28  & 48  \\
                    different lang & 18  & 0\\      
                    incorrect answer & 22 & 36\\
                    annotation error & 22 & 12 \\
                    underspecified q &  10 & 4\\
                    \bottomrule
                \end{tabular}
                \captionof{table}{\small Error categories (\%) on 50 errors sampled from Japanese (Ja) and Spanish (Es) data.} \label{tab:error_analysis}
            }
        }
    \end{minipage}
\end{floatrow}
\end{figure}

\vspace{-.2cm}\paragraph{Breakdown of reference languages.}
\Fref{fig:heatmap} breaks down retrieved reference languages for each target language. 
Our multilingual retrieval model often retrieves documents from the target language (if its reference documents are available), English, or its typologically similar languages. 
For example, mDPR often retrieves Spanish passages for Portuguese questions and Japanese passages for Chinese questions; while they are considered phylogenetically distant, Japanese and Chinese overlap in script. 

 To further evaluate this, we  conduct a controlled experiment: we remove Spanish, Swedish and Indonesian document embeddings and evaluate \mora on related languages: Danish, Portuguese and Malay. We observe performance drops of 1.0 in Danish, 0.6 in Portuguese, and 3.4 F1 points in Malay. 
This illustrates that while \cora allows for retrieval from any language in principle (\textit{many-to-many}), cross-lingual retrieval from closer languages with more language resources is particularly helpful.

\vspace{-.2cm}\paragraph{Error analysis and qualitative examples.}
\Tref{tab:error_analysis} analyzes errors from \mora by manually inspecting 50 Japanese and Spanish wrong predictions from MKQA.
We observe six major error categories: (a) retrieval errors, (b) generating correct answers in a different language (different lang), (c) incorrect answer generation (incorrect answer), (d) answer annotation errors (e.g., a correct alias isn't covered by gold answers, or Wikipedia information is inconsistent with English.), and (e) ambiguous or underspecified questions such as ``who won X \underline{this year}'' (underspecified q).
The table shows that both in Japanese and Spanish, the retrieval errors are dominant. 
In Japanese, \mora often generates correct answers in English, not in Japanese (different lang).

\begin{figure*}[t!]
\vspace{-0.5em}
  \includegraphics[width=\linewidth]{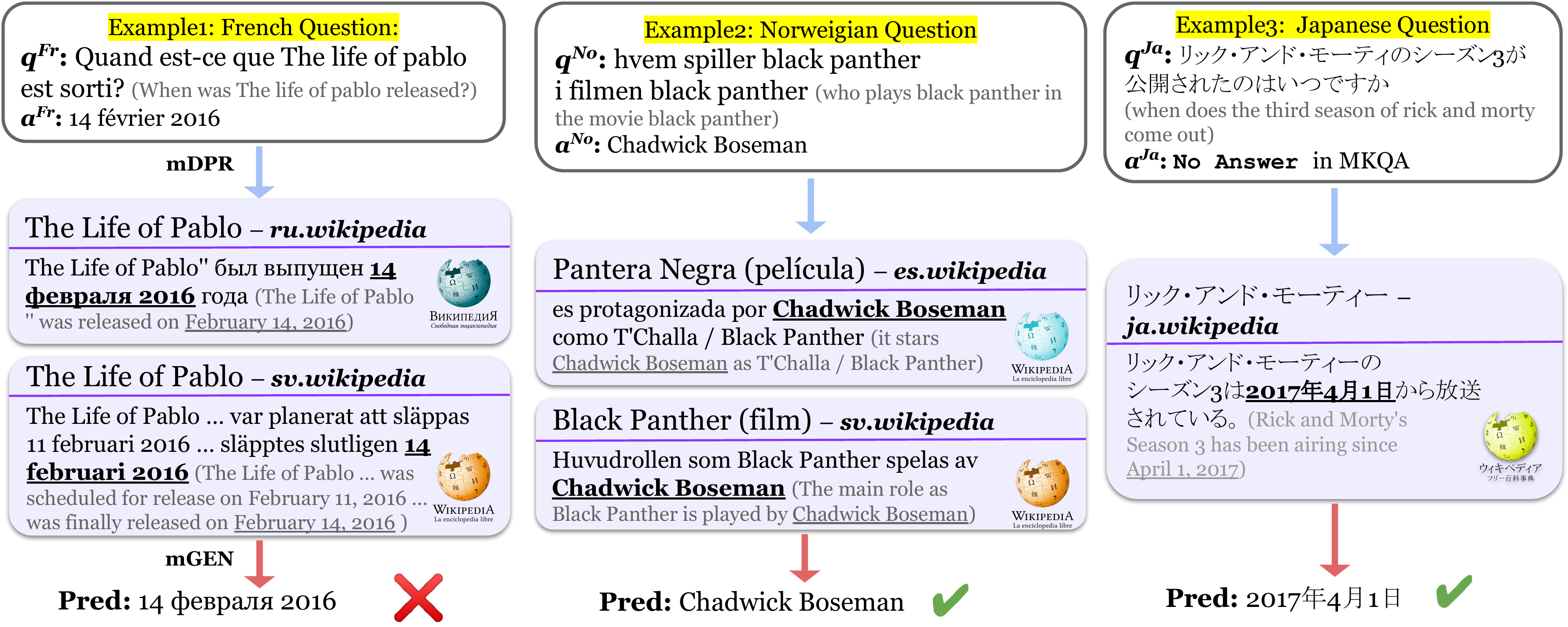}
  \caption{Cross-lingual retrieval and generation examples for three MKQA questions.
}
\vspace{-0.5em}
  \label{img:examples}
\end{figure*}

\Fref{img:examples} shows some qualitative examples. 
The first example shows an error in (b): mGEN is generating an answer in Russian, not in French though the answer itself is correct.
This type of error happens especially when retrieved passages are in languages other than the target and English.  
\vspace{-.2cm}\paragraph{Human evaluation on cross-lingual retrieval results. }
To observe how cross-lingual retrieval between distant languages is actually helping, we sample 25 Norwegian questions for which Spanish passages are included among the top 10 retrieved results.
As seen in \Fref{fig:heatmap}, \mora retrieves Spanish (es) passages for 6.8\% of the Norwegian (no) questions.
A Spanish speaker judges if the retrieved Spanish passages actually answer the given Norwegian questions.\footnote{During evaluation, we provide the original English questions from MKQA. }
We found that in 96\% of the cases, the retrieved Spanish passages are relevant in answering the question. One such example is presented in \Fref{img:examples} (the second example). 

\vspace{-.2cm}\paragraph{Human analysis on unanswerable questions.}
\mora retrieves passages from a larger multilingual document collection than the original human annotations.
Thus, \cora may further improve the answer coverage over the original human annotations. 
MKQA includes questions that are marked as unanswerable by native English speakers given English knowledge sources.
We sample 400 unanswerable Japanese questions whose top one retrieved passage is from a non-English Wikipedia article. 
Among these, 329 unanswerable questions are underspecified (also discussed in \citealp{asai2020unanswerable}).
For 17 out of the 71 remaining questions, the answers predicted by \mora are correct. 
This finding indicates the significance of cross-lingual retrieval and potential room for improvement in annotating multilingual open QA datasets.
The third example in \Fref{img:examples} shows one of these cases.
\section{Related Work and Broader Impacts}
\label{sec:related_work}
\noindent{\bf English and non-English open QA.}
Despite the rapid progress in open QA~\citep{chen2017reading,karpukhin2020dense}, most prior work has been exclusively on English~\citep{lewis2020retrieval,izacard2020leveraging}.
Several prior attempts to build multilingual open QA systems often rely on machine translation or language-specific retrieval models~\citep{ture-boschee-2016-learning, xorqa}. 
\cite{lewis2020retrieval} and \cite{guu2020realm} introduce a similar \textit{retrieve-then-generate}.
\cite{izacard2020distilling} introduce an iterative training framework that uses attention weights from a generator model as a proxy for text relevance scores.
\cite{tran2020cross} introduce CRISS, a self-supervised pre-training approach consisting of a parallel sentence mining module and a sequence-to-sequence model, which are trained iteratively. 
Several recent work such as ~\citet{xiong2021approximate} improves DPR by mining and learning with hard examples. 
Our work is the first work that introduces a unified multilingual system for {\it many-to-many} open QA, which is a challenging task requiring massive-scale cross-lingual retrieval and has not been addressed in prior work. 
We introduce an iterative training and data mining approach guided by filtering from an answer generation model to automatically extend annotated data available only in high-resource languages to low-resource. 
This approach contributes to significant performance improvements in languages without annotated training data. 

\noindent{\bf Many-languages-one models.}
Several recent work introduces single multilingual models for many languages using pre-trained multilingual models such as mBERT or mT5 in many NLP tasks (e.g., entity linking: \citealp{botha-etal-2020-entity, de2021multilingual}; semantic role labeling: \citealp{mulcaire-etal-2019-polyglot, lyu-etal-2019-semantic, fei-etal-2020-cross}; syntactic parsing: \citealp{mulcaire-etal-2019-low, kondratyuk-straka-2019-75}). 
This work conducts the first large-scale study of a unified multilingual open QA model across many languages and achieves state-of-the-art performance in 26 typologically diverse languages.

\noindent{\bf Synthetic data creation for machine reading comprehension.}
\cite{alberti-etal-2019-synthetic} introduce a method of generating synthetic machine reading comprehension data by automatically generating questions and filtering them out by a trained machine reading comprehension model.
Several studies augment multilingual machine reading comprehension training data by generating new question-answer pairs from randomly sampled non-English Wikipedia paragraphs \citep{riabi2020synthetic,shakeri2020multilingual}. 
This work focuses on multilingual open QA, which involves not only machine reading comprehension but also cross-lingual retrieval. A similar augmentation method for machine reading comprehension can be applied to further improve the answer generation component in \cora.

\noindent{\bf Societal impacts.}
Our code and data are publicly available. 
\cora can perform open QA in unseen languages and can benefit society in building QA systems for low-resource languages, hence enabling research in that direction. 
Unlike previous models, \cora removes the necessity of external black-box APIs, and thus we can examine and address wrong answers due to model errors or misinformation present on Wikipedia.
This would help us mitigate the potential negative impact from \cora or its subsequent models outputting a wrong answer when it is used by people who seek information.
\section{Conclusion}
To address the information needs of many non-English speakers, a QA system has to conduct cross-lingual passage retrieval and answer generation. 
This work presents \mora, a unified multilingual {\it many-to-many} open QA model that retrieves multilingual passages in many different languages and generates answers in target languages.
\mora does not require language-specific translation or retrieval components and can even answer questions in unseen, new languages. 
We conduct extensive experiments on two multilingual open QA datasets across 28 languages, 26 of which \mora advances the state of the art on, outperforming competitive models by up to 23 F1 points. 
Our extensive analysis and manual evaluation reveal that \mora effectively retrieves semantically relevant passages beyond language boundaries, and can even find answers to the questions that were previously considered unanswerable due to lack of sufficient evidence in annotation languages (e.g., English).  
Nonetheless, our experimental results show that the retrieval component still struggles to find relevant passages for queries in some unseen languages.
Our analysis also showed that \cora sometimes fails to generate an answer in the target language.
In future work, we aim to address these issues to further improve the performance and scale our framework to even more languages.  
\section*{Acknowledgement}
This research was supported by NSF IIS-2044660, ONR N00014-18-1-2826, gifts from Google,
the Allen Distinguished Investigator Award, the
Sloan Fellowship, and the Nakajima Foundation
Fellowship. We thank anonymous reviewers, area chairs, Eunsol Choi, Sewon Min, David Wadden, and the members of the
UW NLP group for their insightful feedback on this paper, and Gabriel Ilharco for his
help on human analysis. 
\medskip

\bibliographystyle{acl_natbib}
\bibliography{custom}

\clearpage
\appendix

\section*{Appendix}
\section{Details of Modeling}
\subsection{Input format}

\paragraph{Passage representations.}
To create a passage representation, the passage title and text are concatenated (\texttt{[CLS]} \textit{title} \texttt{[SEP]} \textit{passage} \texttt{[SEP]}), following common practice \citep{karpukhin2020dense}. 
We retrieve top 10 passages and use them as input to mGEN. 

\paragraph{Generator input.}
The input to the generator is a concatenation of $q^L$ and $\mathcal{P}^{multi}$. As described in \Sref{sec:method_reader}, we append a language tag that represents $L$ to $q^L$.   
For each passage, we prepend the retrieved ranks and the original Wikipedia article titles and concatenate them to form a input paragraph sequence. 
We differentiate those paragraphs from the question using special tokens (\texttt{<P>} vs.\ \texttt{<Q>}). 
Finally, the concatenated passages are appended to $q^L$ and the language tag. Below is an example input:

\textit{<Q>: \begin{CJK}{UTF8}{min}ロンポールの学部時代の専攻は何？\end{CJK}[ja] <P>:<0:\begin{CJK}{UTF8}{min}ロン・ポール\end{CJK}>\begin{CJK}{UTF8}{min}ロナルド・アーネスト・ポール (英語: Ronald Ernest ``Ron'' Paul、1935年8月20日 - ) は、アメリカ合衆国の元政治家。共和党所属でテキサス州選出の元連邦下院議員であった\end{CJK} <1: Ron Paul> Paul went to Gettysburg College, where he was a member of the Lambda Chi Alpha fraternity. He graduated with a B.S. degree in {\bf Biology} in 1957.}

As in the case of machine translation, we found that the language code does not need to be specified during inference as our model learns the question language automatically. Yet, we found that training with language codes is particularly useful to augment training data for $L_{target}$ without any question data in $L_{target}$. 
In particular, given questions from existing datasets in $L_{source}$ and entities names in $L_{target}$ corresponding to the original answers in $L_{source}$, our generator learns to generate answers in $L_{target}$ from the language code even when questions themselves are written in $L_{source}$. Please see the details of training mGEN with synthetic data in the next section. 

\subsection{Details of the Data Mining Process}
\label{app_sec:data_mining}
\paragraph{Synthetic data for mGEN.}
To train mGEN to generate answers in languages that are not covered by annotated data or our reference sources, we augment  English QA data ($q^{En}, a^{En})$ from Natural Questions~\citep{kwiatkowski2019natural}.
We first use an English DPR model to retrieve $\mathcal{P}^{En}$ for each $q^{En}$. 
Then, we automatically translate $a^{En}$ to a target language $L$ using Wikipedia language links.
We use Media Wiki API,\footnote{\url{https://www.wikidata.org/w/api.php}.} and form new mGEN training data $(q^{En}, a^{L}, \mathcal{P}^{En})$. 
Although the questions and passages are all written in English, our model knows in which language it should answer from the language code appended to the question. 
We limit the target languages for this augmentation process to Arabic, Finnish, Japanese, Korean, Russian, Spanish, Swedish, Hebrew, Thai, Danish, French, Italian, Dutch, Polish, and Portuguese. 
Interestingly, just adding this language code effectively changes the outputs as shown in \Tref{tab:sync_gen_output}.
Although we could create at most 15 synthetic data for each$(q^{En}, a^{En}, \mathcal{P}^{En})$, we sample at most 10 languages from the 15 languages to avoid overfitting. 
We further subsample 50\% of the synthetically generated questions. 
Those synthetically generate data is introduced after training mGEN for 3 epochs to avoid overfitting. 
\begin{table}[h]
\small
                \begin{tabular}{l|c|c}
                    \toprule
                    input question & output & gold answers \\
                    \midrule
                    who is the actor that plays the good doctor [ja] & \begin{CJK}{UTF8}{min}フレッド・ハイモア\end{CJK} &  \begin{CJK}{UTF8}{min}フレディ・ハイモア\end{CJK} \\ 
                    who is the actor that plays the good doctor [ko] & \begin{CJK}{UTF8}{mj}프레디 하이모어\end{CJK} &  \begin{CJK}{UTF8}{mj}프레디 하이모어\end{CJK} \\ 
                    who is the actor that plays the good doctor [it]  & Freddie Highmore & Freddie Highmore \\
                    \bottomrule
                \end{tabular}
                \captionof{table}{Examples of mGEN outputs with varying language codes.} \label{tab:sync_gen_output}
\end{table}

\section{Details of Experiments}
\subsection{Details of the knowledge source language selection.}
In addition to the English Wikipedia embeddings, we encode all of the passages from the Wikipedias of all of the ten languages included in \tydixor or \tydi. 
Adding all of the languages available in Wikipedia to our document collection would significantly increase the index size and slow down inference.
Therefore, we add the languages among the 26 MKQA languages that satisfy the following criteria: (i) a language is included in \tydixor or \tydi, (ii) a language uses non-Latin script and has the largest number of the Wikipedia articles among the languages in the same language family branch (e.g., Thai), or (iii) a language uses Latin script and has more than 1.5 million articles as of May 2021.\footnote{\url{https://en.wikipedia.org/wiki/List_of_Wikipedias}.}

\subsection{Details of Wikipedia statistics}
\label{app_sec:wikipedia_stat}
For our multilingual retriever, we split each article into 100-token chunks \citep{karpukhin2020dense}, while BM25 first splits Wikipedia articles into the pre-defined paragraph units.
We also filter out the short articles with fewer than $k$ (i.e., $k=20$ in this work) tokens, following common techniques in open QA~\citep{min2021neurips} in $\mathbf{C}^{multi}$.
As a result, we add more than 43.6 million articles across the languages.
The original passage text file is 29GB, and the total index size is around 129 GB. 
 \begin{table}
 \vspace{-10pt}
\begin{tabular}{l||r|r}
\toprule
\textbf{language} & \textbf{ The number of articles} & \textbf{The number of passages}  \\
\midrule
English &6,297,085 & 18,003,200\\
Arabic & 664,693  & 1,304,828 \\
Finnish & 451,338 & 886,595 \\
Japanese &	1,268,148 & 5,116,905 \\
Korean & 441,316 & 638,864 \\
Russian &1,522,499 & 4,545,635 \\
Bengali & 64,556 & 179,936 \\
Telugu & 70,356 & 274,230 \\
Indonesian &452,304 & 820,572 \\
Thai & 129,122 & 520,139 \\
Hebrew & 237,836 & 1,045,255 \\
Swedish & 3,758,071  & 4,525,695 \\
Spanish & 1,453,732& 5,738,484 \\
\bottomrule
\end{tabular}
\caption{Statistics of the Wikipedia data.
}
\label{table:wikipedia_stat}
\end{table}

\subsection{Licence, ethical considerations and data splits of \tydixor and MKQA}
\paragraph{Licence.}
Both two datasets are under the MIT licence.
The dataset can be downloaded from their official repositories.\footnote{\url{https://github.com/apple/ml-mkqa} for MKQA; \url{https://github.com/AkariAsai/XORQA} for \tydixor.}

\paragraph{Potential risk of offensive or personally identifiable information.}
MKQA~\citep{mkqa} questions are originally from the Natural Questions data~\citep{kwiatkowski2019natural}. The questions are anonymized Google Search queries, and we expect that those questions are not personally identifiable. 
The Natural Questions authors conduct several procedures to filter out noisy questions, and we expect that the questions do not contain offensive or inappropriate content. 

Likewise, \tydixor questions are from \tydi, where questions are written by their in-house annotates who have native proficiency in the target languages.
The \tydi authors trained those in-house annotators and asked them to write questions that they are interested in given short prompts. We expect that the resulting questions are not personally identifiable and have no risk of offensive information. 

\paragraph{Data splits.}
MKQA does not have any train data, and we use the questions with answer annotations for evaluation, removing 1,427 \texttt{unanswerable} questions and 1,815 \texttt{long\_answer} questions.
Consequently, MKQA evaluation data has 6,758 questions for each target language.
Note that the \texttt{unanswerable} and \texttt{long\_answer} type information is provided in the original MKQA dataset, and we do not conduct any manual data filtering. 
For the ablation or controlled experiments, we randomly sample 350 questions from the 6,758 questions with short answers due to our computational constraints.   

We use the train, dev and test data splits from the original \tydixor (full) data.

\subsection{Language family, branch and script type Information of the languages}
\Tref{table:lang_list} provides a full list of the 28 languages and their language family, branch and script type information. 
The target languages are typologically diverse; 12 of them use their own script system, which makes answer generation in those languages harder than in the languages with Latin script. 
\begin{table*}[t!]
\addtolength{\tabcolsep}{-2pt}
\small
    \centering
    \begin{adjustbox}{width=1.0\textwidth}
    \begin{tabular}{l|ccccc}
\toprule
 &  {\bf Ar} &  {\bf Bn} & {\bf Da} & {\bf De} & {\bf En}  \\\midrule
name & Arabic & Bengali & Danish & German & English \\
family & Afro-Asiatic & Indo-European &Indo-European & Indo-European & Indo-European \\
branch & Semitic  & Indo-Iranian & Germanic & Germanic & Germanic  \\
script & Arab  & Beng  & Latn &Latn  & Latn  \\
 \bottomrule
 & {\bf Es} & {\bf Fi} & {\bf Fr}  & {\bf He} & {\bf Hu}  \\\midrule
name & Spanish & Finnish & French & Hebrew & Hungarian \\
family & Indo-European&  Uralic  &  Indo-European & Afro-Asiatic  & Uralic \\
 branch & Italic & Finic & Italic &  Semitic & Finno-Ugric \\
 script & Latn & Latn & Latn & Hebr & Latn \\ 
 \bottomrule
&  {\bf It}   & {\bf Ja} & {\bf Ko} &   {\bf Km}  & {\bf Ms} \\\midrule
name  &  Italian & Japanese & Korean & Khmer  & Malay \\ 
family &Indo-European& Japonic & Koreanic & 	
Austroasiatic & Austronesian \\
 branch & Italic & Japanese & Korean & Proto-Mon-Khmer & Malayo-Polynesian 
  \\
 script &  Latn & Jpan & Hang & Khmr &Latn  \\
  \bottomrule
  & {\bf Nl} & {\bf No} & {\bf Pl} & {\bf Pt} & {\bf Ru} \\\midrule
name& Dutch & Norwegian & Polish & Portuguese & Russian \\ 
family  & Indo-European & Indo-European & Indo-European & Indo-European & Indo-European   \\
 branch &  Germanic  & Germanic  &  Balto-Slavic & Italic & Balto-Slavic \\
 script & Latn & Latn & Latn & Latn & Cyrl \\
  \bottomrule
  & {\bf Sv}  & {\bf Te}  & {\bf Th}  & {\bf Tr} & {\bf Vi} \\ \midrule
name  & Swedish &  Telugu & Thai & Turkish & Vietnamese \\ 
family & Indo-European &  Dravidian & Kra–Dai & Altaic &  Austroasiatic  \\
 branch &  Germanic &  South-Centra & Tai &Turkic &Vietic  \\
 script  & Latn &  Telu & Thai &Latn & Latn \\
 \bottomrule
 & {\bf Zh-cn} & {\bf Zh-hk} &   {\bf Zh-tw} \\\midrule
name  & Chinese (Simplified) & Chinese (Hong Kong) & Chinese (Traditional)  \\ 
family &  Sino-Tibetan & Sino-Tibetan & Sino-Tibetan \\
 branch  & Chinese & Chinese & Chinese  \\
 script  & Hans/Hant& Hant   & Hant \\
 \bottomrule
    \end{tabular}
    \end{adjustbox}
    \caption{List of 28 language we test in this work. The script is based on ISO 15924.
    }
    \label{table:lang_list}
\end{table*}

\subsection{Hyperparameters of \mora}
\label{app_sec:hyper_params}
\paragraph{mDPR.}
We first fine-tune mDPR on the Natural Questions data using the training data file released by DPR authors.\footnote{\url{https://github.com/facebookresearch/DPR}.}
We filter out questions that are used to create MKQA evaluation data by comparing the input questions. 
We use the same hyperparameters as in the original DPR~\citep{karpukhin2020dense}. 
We then perform fine-tuning on \tydi and \tydixor's gold paragraph data, initializing the checkpoint that achieves the best performance on the development data.
We fine-tune the model for 40 epochs and use the checkpoint that produces the best retrieval performance on \tydixor's development data. 
We use 8 GPUs with 24G RAM, and the total batch size is 128.
We empirically found that using the updated query encoder hurt the retrieval performance on MKQA, while in \tydixor we observe a marginal performance drop. 
Therefore, at inference, we continue using the query encoder trained on the initial data, while we use the updated passage encoder to encode $\mathbf{C}^{multi}$.

\paragraph{mGEN.}
The full list of the hyperparameters are in \Tref{tab:mgen_hyperparam}.
We first train our mGEN using the initial data for 15 epochs and use the checkpoint that gives the highest development score. 
We use Adam~\citep{kingma2014adam} for optimization. 
We subsequently apply iterative training.
During our $t$th iterative step, we use the best checkpoint so far to label new positive and negative passages, which will then be used to fine-tune mDPR at the next iteration. After we finish mGEN training at the $t$-th iteration, we use the best checkpoint for the final evaluation without performing additional data mining. 
We use our internal cluster to run all of the mGEN related training.
We use 8 GPUs with 24G RAM, and the total batch size is 32.

\paragraph{Inference.}
During inference, we first retrieve top 15 passages using mDPR, and then feed the questions and concatenated passages into the mGEN model, with language tags.
We use the same checkpoints and encoded embeddings for MKQA and \tydixor. 
There are minor differences in the gold answer format in MKQA and \tydixor due to different annotation methods (e.g., translate English answers by Wikidata vs.\ use answers extracted from the target language Wikipedias). 
One may fine-tune different models using different subsets of training examples (e.g., MKQA can benefit from more NQ-based synthetic training data as the questions are originally from NQ).
In this work, we focus on building a unified QA system across languages and datasets, and thus use the same checkpoints for all of the experiments. 

 \begin{table}
 \vspace{-10pt}
\begin{tabular}{l|r}
\toprule
\textbf{hyperparameter} &  \\
\midrule
max source length & 1,000 \\
max target length & 25 \\
batch size (per GPU) & 2 \\
label smoothing & 0.1 \\
dropout & 0.1 \\
warmup steps & 500 \\
learning rate & 3e-5 \\
weight decay & 0.001 \\
adam $epsilon$ & 1e-8 \\
max grad norm & 0.1 \\
gradient accumulation steps & 2 \\
\bottomrule
\end{tabular}
\caption{Hyperparameters of mGEN.
}
\label{tab:mgen_hyperparam}
\end{table}

\subsection{Details of translate-test baseline}
We first translate the MKQA and \tydixor questions from various languages to English, use DPR to retrieve the answers from the knowledge source, use the reader to extract an answer, and then translate the answer back to its original language.

\paragraph{Details of translation models.}
Our translation models (to English and from English) are the pre-trained MarianMT (\citealp{mariannmt}) style OPUS-MT (\citealp{TiedemannThottingal:EAMT2020}) models available in Transformers library (\citealp{wolf-etal-2020-transformers}) that are trained on the OPUS corpus \citep{tiedemann-nygaard-2004-opus}.
Since there is no MarianMT pre-trained OPUS-MT model from English to Korean on Transformers, we use the pre-trained base-sized autoregressive transformers model provided by the authors of \tydixor.\footnote{\url{https://github.com/jungokasai/XOR_QA_MTPipeline}.}

Some of the newer OPUS-MT models require a prefix of the target language before each sentence of the source language (English here) when translating English answers back to the question’s original language, which is usually the ISO 639-3 language code. 
For example, if we want to translate {\it Ron Paul} from English to Arabic, we concatenate the prefix ``\texttt{>>ara<<}'' and the original sentence together to specify the target language to be Arabic since the \texttt{opus-mt-en-ar} \footnote{\url{https://huggingface.co/Helsinki-NLP/opus-mt-en-ar}.} model supports multiple target languages. Then, we feed the concatenated result \textit{``}\texttt{>>ara<<} \textit{Ron Paul''} into the translation model and get the translation.

Such prefixes and the models we use for each language are listed in Table \ref{table:trans-model}. 

\begin{table*}[t!]
    \small
    \centering
    \begin{adjustbox}{width=1.0\textwidth}
    \begin{tabular}{l||cccc}
    \toprule
    & {\bf Ar} & {\bf Da} & {\bf De} & {\bf Es} \\\hline
    To-English Model & \texttt{opus-mt-ar-en}  & \texttt{opus-mt-da-en} & \texttt{opus-mt-de-en} & \texttt{opus-mt-es-en}\\
    From-English Prefix & \texttt{>>ara<<} & N/A & N/A & N/A \\
    From-English Model & \texttt{opus-mt-en-ar} & \texttt{opus-mt-en-da} & \texttt{opus-mt-en-de} & \texttt{opus-mt-en-es} \\
    \bottomrule
     & {\bf Fi} & {\bf Fr} & {\bf He} & {\bf Hu} \\\hline
     To-English Model & \texttt{opus-mt-fi-en}  & \texttt{opus-mt-fr-en} & \texttt{opus-mt-afa-en} & \texttt{opus-mt-hu-en}\\
    From-English Prefix & N/A  & N/A & \texttt{>>heb<<} & N/A \\
    From-English Model & \texttt{opus-mt-en-fi} & \texttt{opus-mt-en-fr} & \texttt{opus-mt-en-afa} & \texttt{opus-mt-en-hu} \\
    \bottomrule
    & {\bf It} & {\bf Ja} & {\bf Ko} & {\bf Km} \\\hline
     To-English Model & \texttt{opus-mt-it-en}  & \texttt{opus-mt-ja-en} & \texttt{opus-mt-ko-en} & \texttt{opus-mt-mul-en}\\
    From-English Prefix & N/A  & N/A & N/A & \texttt{>>khm\_Latn<<} \\
    From-English Model & \texttt{opus-mt-en-it} & \texttt{opus-mt-en-jap} & N/A & \texttt{opus-mt-en-mul} \\
    \bottomrule
    & {\bf Ms} & {\bf Nl} & {\bf No} & {\bf Pl} \\\hline
     To-English Model & \texttt{opus-mt-mul-en}  & \texttt{opus-mt-nl-en} & \texttt{opus-mt-gem-en} & \texttt{opus-mt-pl-en}\\
    From-English Prefix & \texttt{>>zsm\_Latn<<}  & N/A & \texttt{>>nno<<} & \texttt{>>pol<<} \\
    From-English Model & \texttt{opus-mt-en-mul} & \texttt{opus-mt-en-nl} & \texttt{opus-mt-en-gem}  & \texttt{opus-mt-en-sla} \\
    \bottomrule
    & {\bf Pt} & {\bf Sv} & {\bf Th} & {\bf Tr} \\\hline
     To-English Model & \texttt{opus-mt-ROMANCE-en}  & \texttt{opus-mt-sv-en} & \texttt{opus-mt-th-en} & \texttt{opus-mt-tr-en}\\
    From-English Prefix & \texttt{>>pt<<}  & N/A & \texttt{>>tha<<} & \texttt{>>tur<<} \\
    From-English Model & \texttt{opus-mt-en-ROMANCE} & \texttt{opus-mt-en-sv} & \texttt{opus-mt-en-mul}  & \texttt{opus-mt-en-trk} \\
    \bottomrule
    & {\bf Vi} & {\bf Zh-cn} & {\bf Zh-hk} & {\bf Zh-tw} \\\hline
     To-English Model & \texttt{opus-mt-vi-en}  & \texttt{opus-mt-zh-en} & \texttt{opus-mt-zh-en} & \texttt{opus-mt-zh-en}\\
    From-English Prefix & \texttt{>>vie<<}  & \texttt{>>cmn<<} & \texttt{>>yue\_Hant<<} & \texttt{>>cmn\_Hant<<} \\
    From-English Model & \texttt{opus-mt-en-vi} & \texttt{opus-mt-en-zh} & \texttt{opus-mt-en-zh}  & \texttt{opus-mt-en-zh} \\
    \bottomrule
    \end{tabular}
    \end{adjustbox}
    \caption{Translation models and prefixes used for the translate-test baseline.}
    \label{table:trans-model}
\end{table*}

\paragraph{Details of English DPR model.}
For the English DPR model, we use the trained retriever and reader models from \tydixor, which can be downloaded from their official website.\footnote{\url{https://github.com/AkariAsai/XORQA/tree/main/baselines}.}

\subsection{Details of BM25 baseline}
We use the February 2019 Wikipedia dumps as the knowledge source and retrieval corpus for our BM25 baseline. 
We first use wikiextractor to preprocess the Wikipedia documents, and then use Pyserini (\citealp{lin2021pyserini}), which relies on Apache Lucene 8.0.0\footnote{\url{https://lucene.apache.org/core/8_0_0/index.html}{\url{https://lucene.apache.org/core/8\_0\_0/index.html}}.} to index the documents and retrieve the BM25 results for \tydixor and MKQA questions. 
We use 2 paragraphs as one basic unit of retrieval where paragraphs are separated by \texttt{`\textbackslash n'}.
We retrieve the top 10 units that have the highest BM25 score. 
After we retrieve top units, we concatenate those paragraphs and feed them into a bert-base-multilingual-uncased extractive QA model that predicts start and end positions. 
The final answers are determined as the span with the highest joint probabilities. 

French, Hebrew, Khmer, Malay, Polish, Vietnamese and Chinese (Hong Kong) are either not supported by Apache Lucene or missing from the Wikipedia dumps, and therefore are not included in the final results.

\subsection{Details of MT+Mono baseline}
We normalize the predicted probabilities from the the BM25 (monolingual) baseline so that the score will be between 0 to 1. We output the monolingual baseline's answer when the probability is higher than a threshold; otherwise, we output translated answers from the translate-test baseline. 
In this work, we set the threshold to 0.5 given the results on the \tydixor development set. 

\subsection{Details of the Closed-book Baseline}
Instead of training a sequence-to-sequence model from scratch using question and answer only data as in \cite{roberts2020much}, we use the same mGEN model as in \cora, and only at inference time do we skip retrieval. We also tested an mt5-base based sequence-to-sequence model that is trained to generate answers from questions only, but this model underperformed the inference-only model on the \tydixor development set. 

\section{Additional Results}
\subsection{Results on \tydixor Development Set}
\Tref{tab:xor_dev} shows the results on the \tydixor development set.
We clearly outperform the previous state-of-the-art model, as well as the competitive baselines, by a large margin across target languages. 
The scores on \xorfull development set are significantly higher than the \xorfull test set presented in \Tref{table:main_results}. We have found that the proportions of the questions where answers can be extracted from the target languages' Wikipedia are significantly higher than in \xorfull test set. 
Our \mora framework improves the performance on those ``in-language'' subsets of \xorfull and get a large performance jump on the test set. 
\begin{table*}[t!]
\addtolength{\tabcolsep}{-0.5pt}
\small
    \centering
    \begin{tabular}{l|ccccccc|ccc}
\toprule
 Models & \multicolumn{7}{c}{Target Language $L_i$ F1} & \multicolumn{3}{|c}{Macro Average} \\
 & {\bf Ar} & {\bf Bn} & {\bf Fi} & {\bf Ja} & {\bf Ko} & {\bf Ru} & {\bf Te} & {\bf F1} & {\bf EM} & {\bf BLEU}  \\ \midrule
\mora & {\bf 42.9} & 26.9& {\bf 41.4} & {\bf 36.8} & {\bf 30.4} & {\bf 33.8} & {\bf 30.9} & {\bf 34.7} & {\bf 25.8} & {\bf 23.3} \\
GMT+GS &  18.0 & {\bf 29.1} & 13.8 & 5.7 & 15.2 & 14.9 & 15.6 & 16.0 & 9.9 & 14.9  \\
MT+ Mono & 15.8 & 9.6 & 20.5 & 12.2 & 11.4 & 16.0 & 0.5 & 17.3 & 7.5 & 10.7 \\
MT+DPR &  7.2 & 4.3 & 17.0 & 7.9 & 7.1 & 13.6 & 0.5 & 8.2 & 3.8 & 6.8 \\
BM25 & 18.4 & 14.9 & 18.8 & 12.7 & 12.1 & 13.5  & -- & -- & -- & -- \\
\hdashline
Closed-book & 14.0 & 8.1 & 11.8 & 19.1 & 9.3 & 10.5 & 7.6 & 11.5& 8.2 & 4.9 \\
 \bottomrule
    \end{tabular}
    \caption{Performance on \xorfull (development data F1 scores and macro-averaged F1, EM, and BLEU scores). 
    ``GMT+GS'' denotes the previous state-of-the-art model, which combines Google Custom Search in the target language and Google Translate + English DPR for cross-lingual retrieval \citep{xorqa}. Pyserini does not support Telugu.
    }
    \label{tab:xor_dev}
\end{table*}

\subsection{EM Scores on \tydixor and MKQA}
\paragraph{EM scores on \tydixor.}
The EM scores on the \tydixor test data are in \Tref{tab:xor_test_em}.
We significantly outperform all other baselines and previous state-of-the-art models in all languages except for Korean.
We found that in Korean, our model is often penalized because outputs are correct yet generated in English, not Korean.
The state-of-the-art model ensures that the answers are in Korean using Google Translate, which helps the system to get high performance in Korean. 

\paragraph{EM scores on MKQA.}
The EM scores on MKQA test set are shown in Tables~\ref{tab:mkqa_em_scores_seen} and \ref{tab:mkqa_em_scores_unseen}.
\cora outperforms the other baselines by large margins in all of the languages except for Arabic and English. 
Note that EM scores may underestimate the models' ability of open retrieval; generated answers may be correct even if they do not have a matching sub-span in existing documents, existing Wikidata entries, or human translated answers~\citep{xorqa}.

\begin{table*}[t!]
\addtolength{\tabcolsep}{-0.5pt}
\small
    \centering
    \begin{tabular}{l|ccccccc}
\toprule
 Models & \multicolumn{7}{c}{Target Language $L_i$ EM}  \\
 & {\bf Ar} & {\bf Bn} & {\bf Fi} & {\bf Ja} & {\bf Ko} & {\bf Ru} & {\bf Te}  \\ \midrule
\mora & {\bf 38.4} & {\bf 26.6} & {\bf 33.1} & {\bf 30.1} &  18.9 &  {\bf 36.3} & {\bf 34.6} \\
SER & 23.0 & 16.1 & 18.5 & 6.3 & 9.3 & 6.9 & 14.4 \\
GMT+GS & 22.1 & 10.9 & 13.3 & 3.0 & {\bf 20.1} & 11.4 & 9.1   \\
MT+ Mono & 19.1 & 9.0 & 16.7 & 7.1 & 8.3 & 13.1 & 0.5 \\
MT+DPR &  2.5 & 1.5 & 10.4 & 3.3 & 2.9 & 2.5 & 0.5  \\
BM25 & 23.2 & 14.6 & 17.0 & 5.3 & 9.4 & 13.4 & --  \\
\hdashline
Closed-book & 9.6 & 6.7 & 8.8 & 16.8 & 7.8 &15.5 & 1.9 \\
 \bottomrule
    \end{tabular}
    \caption{Performance on \xorfull (test data EM scores). 
    ``GMT+GS'' denotes the previous state-of-the-art model, which combines Google Custom Search in the target language and Google Translate + English DPR for cross-lingual retrieval \citep{xorqa}. Concurrent to our work, ``SER'' is the current state-of-the-art model, Single Encoder Retriever, submitted anonymously on July 14 to the \xorfull leaderboard (\url{https://nlp.cs.washington.edu/xorqa/}). 
    Pyserini does not support Telugu.
    }
    \label{tab:xor_test_em}
\end{table*}
\begin{table*}[t!]
\footnotesize
    \centering
   \begin{tabular}{l||ccccccccccc}
\toprule
  Setting & -- & \multicolumn{6}{c}{Included in \tydixor  } & \multicolumn{4}{|c}{ Seen by mDPR and mGEN } \\\hline
 & Avg. over all $L$. & {\bf En} &  {\bf Ar} & {\bf Fi} & {\bf Ja} & {\bf Ko} & {\bf Ru}  & {\bf Es} & {\bf Sv} & {\bf He} & {\bf Th} \\\hline
 \midrule
\mora  & {\bf 17.2} & 31.2 & 7.7 & \bf 21.8 & \bf 7.4 & \bf 7.3 & \bf 13.8 & \bf 25.8 &\bf  26.0 & \bf 10.7 &\bf 4.8 \\
MT+Mono & 9.6 & \bf 32.1 & 3.9 & 12.6 &  3.8 & 4.0 &  6.5 & 15.2 & 14.4 & 4.4 & 5.0 \\
MT+DPR& 10.2 & 32.1 & \bf 8.5 & 15.0 & 3.6 & 2.7 & 8.6 & 18.6 & 4.4 & 2.8 & 4.6  \\
BM25 &  -- & 12.8 & 3.1 & 6.5 & 2.7 & 3.9 & 4.7 & 9.4 & 6.6 & -- & 2.6 \\\hdashline
Closed& 2.3 &  3.4 & 2.0 & 2.0 & 3.1 & 2.0 & 2.3 & 3.1 & 4.4 & 2.0 & 2.6 \\ 
\bottomrule
    \end{tabular}
    \caption{EM scores on MKQA seen languages over 6.7k questions with short answer annotations.
    }
    \label{tab:mkqa_em_scores_seen}
\end{table*}
\begin{table*}[t!]
\scriptsize
    \centering
 \begin{adjustbox}{width=\textwidth}
   \begin{tabular}{l||ccccccccccccccccc}
\toprule
 Setting & \multicolumn{7}{c}{ Seen in mGEN } & \multicolumn{9}{|c}{ Unseen}\\\hline
&  {\bf Da} & {\bf De} &  {\bf Fr}  & {\bf It}  & {\bf Nl}  & {\bf Pl}  & {\bf Pt}   &  {\bf Hu}  & {\bf Vi}   & {\bf Ms}  &   {\bf Km} & {\bf No} & {\bf Tr} & {\bf cn} & {\bf hk} &  {\bf tw} \\ \midrule
 \midrule
\mora  &  \bf 25.8 & \bf 25.4 & \bf 25.4 & \bf 24.0 & \bf 26.5 & \bf 21.4 & \bf 23.0 & \bf 15.4 & \bf 17.7 & \bf 23.2 & \bf 4.8 & \bf 24.3 & \bf 18.5 & \bf 4.1 & \bf 5.5 &\bf 4.5 \\
MT+Mono& 13.2 & 15.8 & 15.4 & 15.3 & 15.3 & 14.6 &  13.9 & 11.5 & 7.4 & 3.9 & 0.1 & 11.6 & 11.2 & 3.5 & 2.8 & 4.1  \\
MT+DPR&  16.9 & 16.9 & 15.4 & 16.1 & 17.5 & 14.6 & 14.6 & 10.3 &  7.4 & 3.9 & 0.1 & 13.6 & 8.7 & 1.1 & 2.8 & 3.6 \\
BM25 & 5.8 & 8.7 & -- & 9.4 & 8.4 & -- & 8.7 & 4.8 & -- & -- & -- & 5.6 & 5.6 &  2.5 & -- & 2.8 \\\hdashline
Closed& 2.6 & 3.2 & 2.6 & 2.6 & 2.8 & 2.4 & 2.6 & 2.3 & 2.2 & 2.3 & 1.4  & 2.3 &  1.7 & 2.1 & 1.8 & 1.9  \\ 
\bottomrule
    \end{tabular}
     \end{adjustbox}
    \caption{EM scores on MKQA in languages unseen by mDPR and  not included in $\mathbf{C}^{multi}$. }``cn'': ``zh-cn'' (Chinse, simplified). ``hk'': ``zh-hk'' (Chinese, Hong Kong). ``tw'': ``zh-tw'' (Chinese, traditional). 
    \label{tab:mkqa_em_scores_unseen}
\end{table*}

\section{Further Analysis}
\subsection{Visualizing the Document Embedding Spaces}
\begin{figure}%
    \centering
    \subfloat[\centering Document embeddings before cross-lingual training. ]{{\includegraphics[width=6cm]{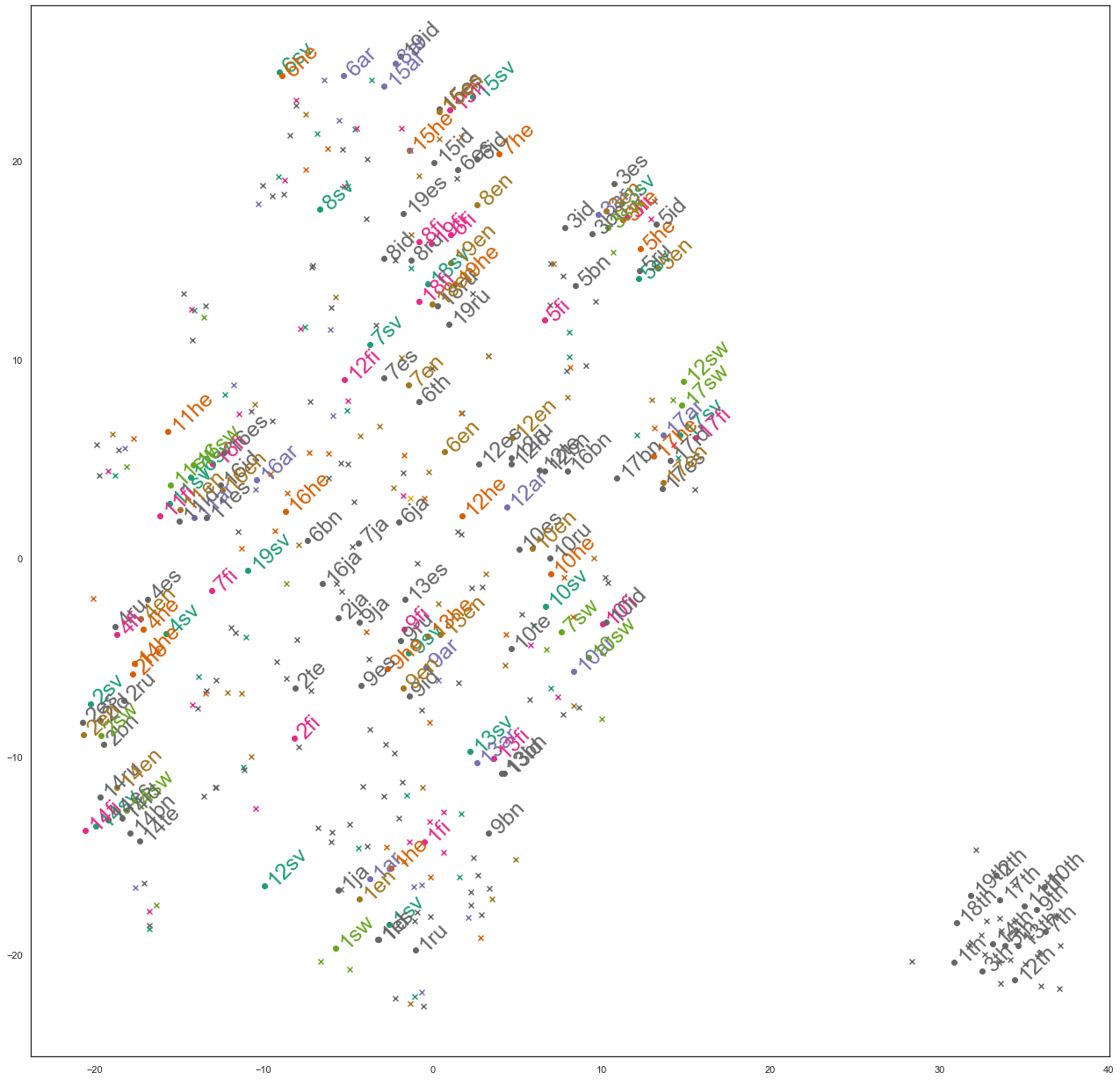} }}%
    \qquad
    \subfloat[\centering Document embeddings after iterative training.]{{\includegraphics[width=6cm]{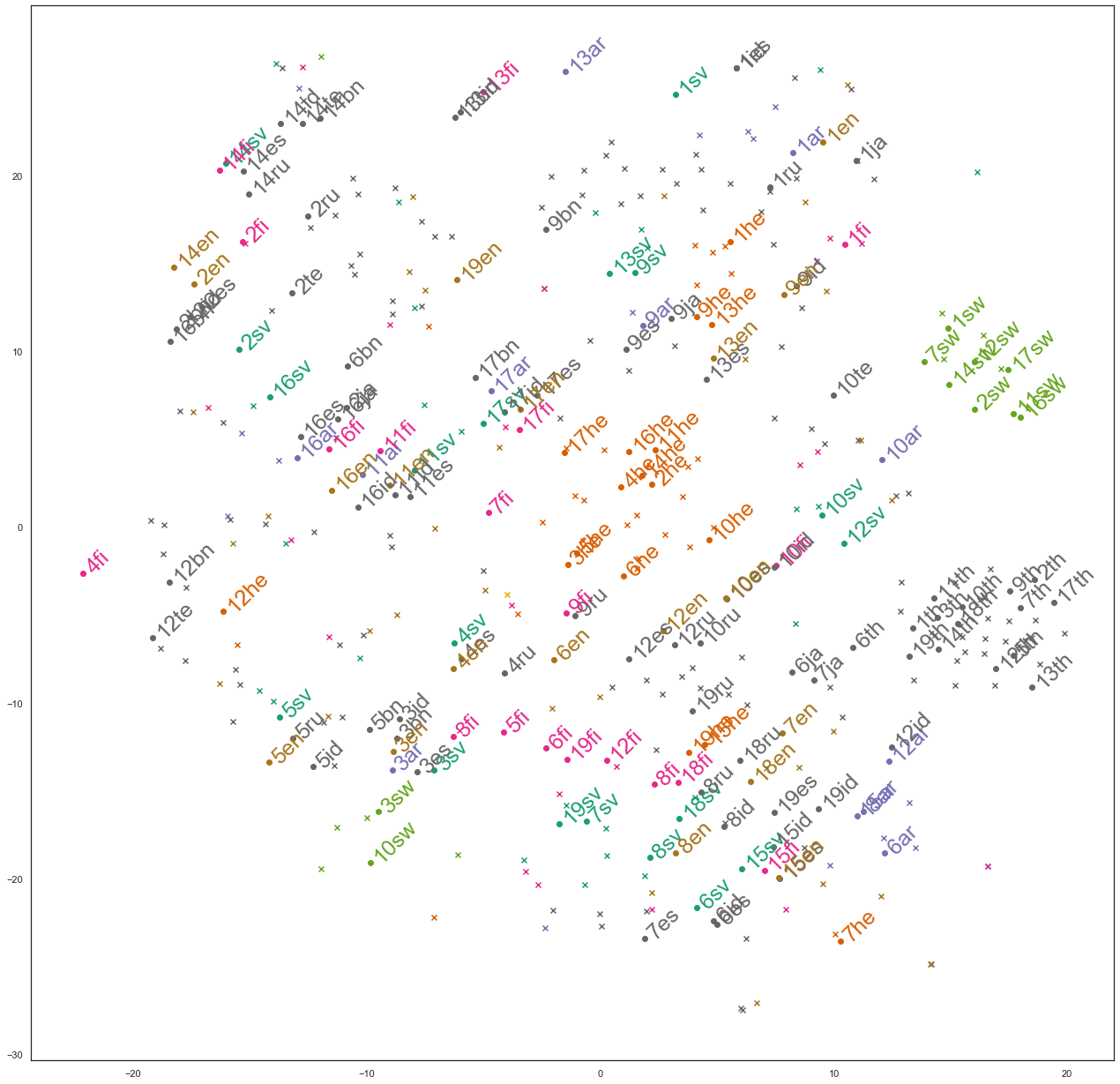} }}%
    \caption{Embeddings before and after cross-lingual training (PCA).}%
    \label{fig:doc_embeddings_example}%
\end{figure}

We plot two dimensional encoded document representations (PCA) for the corresponding articles in \Fref{fig:doc_embeddings_example}. 
The gray dots concentrated in the lower right part in the first figure represent encoded Thai embeddings. As we can see from the plot before cross-lingual training, the Thai document embeddings are far apart from other languages' embeddings. On the other hand, after iterative training (Fig.\ \ref{fig:doc_embeddings_example}(b)), the embeddings from many languages get closer, though we can still see loose clusters of some languages. 

\subsection{Analysis on the Cross-lingual Retrieval}
\paragraph{Spanish paragraphs retrieved for Norwegian questions.}
The Spanish paragraphs retrieved for Norwegian questions are shown in \Tref{tab:example_no_es_paras}.
As we can see, \cora retrieves the Spanish passages relevant to the given Norwegian questions and generate correct answers. 
Although those two languages both belong to the Indo-European family and use Latin script, their typological properties (e.g., syntax and vocabulary) differ significantly.
\begin{table*}
\small
\begin{center}
\begin{tabular}{p{120pt}|p{175pt}|p{40pt}}
\toprule
Query & Paragraph & Gold Answer \\ \midrule
hvem spilte maria magdalena i jesus christ superstar (trans: who played mary magdalene in jesus christ superstar) & Los actores principales de la película eran Ted Neeley en el papel de Jesús, Carl Anderson en el de Judas e {\bf Yvonne Elliman} en el papel de María Magdalena. ({\bf trans}: The main actors in the film were Ted Neeley as Jesus, Carl Anderson as Judas, and Yvonne Elliman as Mary Magdalene. ) & Yvonne Elliman \\ \hline
hvor mange episoder er det i andre sesong av my hero academia (trans: how many episodes are in season two of my hero academia) & El estreno fue 7 de abril del 2018. Contando con un total de 25 episodios igual que la segunda. ({\bf trans}: The premiere was April 7, 2018. With a total of 25 episodes the same as the second.) & 25.0 episodes \\
 \bottomrule
\end{tabular} 
\end{center}
\caption{Cross-lingual retrieval examples between Norweigian questions and Spanish passages that lead to correct answers.}
\label{tab:example_no_es_paras}
\end{table*}

\subsection{Analysis on Errors on MKQA Data.}
\paragraph{Details of the annotation process.}
We randomly sample 50 errors from Spanish and Japanese, and we classify those errors into five categories described in \Sref{sec:analysis}. 
Each sample includes: a question in the target language, a question in English (the original NQ question), the top one passage retrieved by mDPR, an answer generated by mGEN, and gold answers.  
The error analysis is conducted by bilingual or native speakers of Spanish or Japanese.

\paragraph{Error analysis results on Chinese examples.}
We also conduct the same analysis in Chinese to understand the relatively low performance in the three Chinese languages. Among the three Chinese variants, we choose simplified Chinese (Zh-cn). The error analysis is conducted by a native speaker. 
\Tref{tab:error_analysis_ja_es_zh} shows the error analysis results in Japanese, Spanish, and Chinese. Generating answers in different languages is common in Chinese like Japanese. Retrieval errors account for 70\% of the errors in Chinese, which is significantly higher than the proportions in Japanese or Spanish. This can be explained by the fact that we do not include Chinese passages in our $\mathbf{C}^{multi}$, and thus \mora always has to conduct cross-lingual retrieval to get evidence to answer. Retrieving documents cross-lingually is more challenging than retrieving documents monolingually, and future work can improve cross-lingual retrieval particularly between languages that are distant from each other. 

\begin{table}[t!]
    \centering
    \small
                \begin{tabular}{l|ccc}
                    \toprule
                     & Japanese &  Spanish & Chinese (simplified) \\
                    \midrule
                    retrieval error & 28  & 48 & 70  \\
                    different lang & 18  & 0 & 10 \\      
                    incorrect answer & 22 & 36 & 4 \\
                    annotation error & 22 & 12 & 4\\
                    underspecified question &  10 & 4 & 12 \\
                    \bottomrule
                \end{tabular}
                \captionof{table}{\small Error categories (\%) on 50 errors sampled from Japanese,  Spanish and Chinese (simplified) data. Error questions are sampled from the MKQA evaluation data.} \label{tab:error_analysis_ja_es_zh}
\end{table}

\subsection{More Qualitative Examples}
\paragraph{Examples errors in MKQA Japanese questions.}
\Tref{tab:error_examples} shows examples of (b) a generation language error (different lang), (c) incorrect answer generation (incorrect answer), and (d) an answer annotation error (annotation error).
The first example shows an error of different language where generated text is not in the target language.
Such errors are prevalent in Japanese, especially when retrieved passages are written in languages with Latin script. 
Transliteration of foreign words into Japanese is challenging as there are multiple ways to map English words to Japanese type script (i.e., \textit{katakana}). 
Future work can improve those cross-lingual generations between languages with their own type script and the ones with Latin script. 
In the second example, \cora answers the song writers, instead of answering who sings the song.
This shows even state-of-the-art models still exploit certain (spurious) patterns or lexical overlap to the question~\citep{sugawara-etal-2018-makes}.
The final example demonstrates the annotation difficulty of covering all possible answer aliases for multilingual open QA. Although the predicted answer is semantically correct, it's not covered by the gold answer annotations in MKQA. 

We also show questions that are judged as (e) underspecified questions in \Tref{tab:invalid_questions} in this analysis.
The first two examples show a question with temporal dependency and an ambiguous question.
In the final example, we found that the information about the population of Utica, New York is different in English Wikipedia (60,635) and Japanese Wikipedia (62,235), and \cora generates an answer based on the information in Japanese.  

\begin{table*}
\small
\begin{center}
\begin{adjustbox}{width=\textwidth}
\begin{tabular}{p{40pt}|p{75pt}|p{170pt}|p{90pt}}
\toprule
\textbf{error type} & Query & Paragraph & Prediction [gold Answer] \\ \midrule
different lang & \begin{CJK}{UTF8}{min}マルコム in the Middleで父親役は誰でしたか\end{CJK} (\textbf{trans}: who played the dad on malcolm in the middle?) & The series is about a boy named Malcolm (Frankie Muniz), the third-born child in a comically dysfunctional working-class family of four, and later, five boys, the sons of Lois (Jane Kaczmarek) and Hal (Bryan Cranston). &  Bryan Cranston [\begin{CJK}{UTF8}{min}ブライアン・クランストン\end{CJK} (\textbf{trans}:Bryan Cranston)] \\ \hline
incorrect answer & \begin{CJK}{UTF8}{min}「愛はとまらない」を唄っているのは誰ですか\end{CJK} (\textbf{trans}: who sings nothing's gonna stop us now?) & \begin{CJK}{UTF8}{min}愛はとまらないはアルバート・ハモンドとダイアン・ウォーレンの共作による楽曲。アメリカ合衆国のロックバンド、 スターシップにより録音された。\end{CJK} (\textbf{trans}:Nothing's Gonna Stop is a song co-written by Albert Hammond and Diane Warren. Recorded by the American rock band Starship.) & \begin{CJK}{UTF8}{min}アルバート・ハモンドとダイアン・ウォーレン\end{CJK} (\textbf{trans}: Albert Hammond and Diane Warren) [\begin{CJK}{UTF8}{min}スターシップ\end{CJK}, Starship] \\ \hline
annotation error & \begin{CJK}{UTF8}{min}マクドナルドの最初の店舗はどこ？\end{CJK} (\textbf{trans}: where was the very first mcdonald's built?)& \begin{CJK}{UTF8}{min}最初のマクドナルドはアメリカ合衆国・カリフォルニア州サンバーナーディノでマクドナルド兄弟が1940年に始めたものである。\end{CJK} (\textbf{trans}: The first McDonald's was started in 1940 by the McDonald's brothers in San Bernardino, California, United States.) &  \begin{CJK}{UTF8}{min}アメリカ合衆国・カリフォルニア州サンバーナーディノ\end{CJK} [\begin{CJK}{UTF8}{min}アメリカ合衆国, カリフォルニア州, サンバーナーディノ, アメリカ\end{CJK}] \\
 \bottomrule
\end{tabular} 
\end{adjustbox}
\end{center}
\caption{Examples of the Japanese error cases. ``\textbf{trans}'' denotes the English translation.}
\label{tab:error_examples}
\end{table*}

\begin{table*}
\small
\begin{center}
\begin{tabular}{p{80pt}|p{140pt}|p{130pt}}
\toprule
\textbf{Sub type} & Query & Prediction [gold Answer] \\ \midrule
temporal dependency & \begin{CJK}{UTF8}{min}今年のスーパーボウルはどこでありますか\end{CJK} (\textbf{trans}: where is the super bowl being played at \hl{this year})&  \begin{CJK}{UTF8}{min}ルイジアナ州ニューオーリンズ\end{CJK} [\begin{CJK}{UTF8}{min}アトランタ\end{CJK}] \\ \hline
ambiguous questions & \begin{CJK}{UTF8}{min}トワイライトシリーズの本を教えてください\end{CJK} (\textbf{trans}: what are the books in the twilight series) &\begin{CJK}{UTF8}{min}ステファニー・メイヤー\end{CJK} [\begin{CJK}{UTF8}{min}エクリプス/トワイライト・サーガ, エクリプス, ニュームーン\end{CJK}] \\\hline
 inconsistency between Wikipedias & \begin{CJK}{UTF8}{min}ニューヨーク州ユーティカの人口はどのくらいですか。\end{CJK} (\textbf{trans}: what is the population of utica new york) & \begin{CJK}{UTF8}{min}62,235人\end{CJK} [60635] \\
 \bottomrule
\end{tabular} 
\end{center}
\caption{Examples of questions labeled as underspecified questions in our error analysis.}
\label{tab:invalid_questions}
\end{table*}

\paragraph{Unanswerable MKQA questions that \mora could answer.}
\Tref{tab:unanswerable_japanese_mkqa} shows \texttt{unanswerable} Japanese MKQA questions for which \cora can successfully find correct answers from non-English languages' text. 
Although MKQA answers are carefully annotated by crowd workers who extensively search online knowledge sources in English, around 30\% of the questions remain unanswerable. Among the valid unanswerable questions, we found that in about 20\% of the \textit{unanswerable} questions we can find correct answers by retrieving evidence passages in another language (e.g., Japanese, Spanish). 
This indicates the effectiveness of cross-lingual retrieval to improve answer coverage.

\begin{table*}
\small
\begin{center}
\begin{adjustbox}{width=\textwidth}
\begin{tabular}{p{100pt}|p{210pt}|p{70pt}}
\toprule
Query & Paragraph & Prediction \\ \midrule
 \begin{CJK}{UTF8}{min}オレンジ・イズ・ニュー・ブラックはいつ放送される？\end{CJK} (\textbf{trans}:when is orange is the new black on?) & \begin{CJK}{UTF8}{min}『オレンジ・イズ・ニュー・ブラック』は、2013年7月11日よりネットフリックスで配信開始されているアメリカのテレビドラマ \end{CJK} (\textbf{trans}:``Orange is the New Black'' is an American TV drama that has been available on Netflix since July 11, 2013.) & \begin{CJK}{UTF8}{min}2013年7月11日\end{CJK} \\ \hline
\begin{CJK}{UTF8}{min}hulkを演じる役者は誰\end{CJK} (\textbf{trans}: who is the actor that plays the hulk?) & Bruce Banner es un personaje interpretado primero por Edward Norton y actualmente por Mark Ruffalo en la franquicia cinematográfica Marvel Cinematic Universe (MCU) basado en el personaje de Marvel Comics del mismo nombre y conocido comúnmente por su alter ego, Hulk (\textbf{trans}: Bruce Banner is a character played first by Edward Norton and currently by Mark Ruffalo in the Marvel Cinematic Universe (MCU) film franchise based on the Marvel Comics character of the same name and commonly known by his alter ego, Hulk.) & Mark Ruffalo \\\hline
\begin{CJK}{UTF8}{min}現在の火星の気温は？\end{CJK} (\textbf{trans}: what's the temperature on mars right now?) & \begin{CJK}{UTF8}{min}現在の火星の表面での年平均気温は、210K以下であり\end{CJK} (\textbf{trans}:The current average annual temperature on Mars is less than 210K.) & \begin{CJK}{UTF8}{min}210K以下\end{CJK}  \\
 \bottomrule
\end{tabular} 
\end{adjustbox}
\end{center}
\caption{Examples of \texttt{unanswerable} Japanese MKQA questions where \mora successfully finds the correct answers. The answers are validated by the authors of this paper. }
\label{tab:unanswerable_japanese_mkqa}
\end{table*}

\end{document}